\documentclass[3p,times]{elsarticle}

\journal{Artificial Intelligence in Medicine}

\newcommand{\etal}{\textit{et al}.}

\makeatletter
\def\ps@pprintTitle{%
 \let\@oddhead\@empty
 \let\@evenhead\@empty
 \def\@oddfoot{}%
 \let\@evenfoot\@oddfoot}
\makeatother

\usepackage{amsmath}
\usepackage{amssymb}
\usepackage{array}
\usepackage{caption}
\usepackage{color}
\usepackage{float}
\usepackage{graphicx}
\usepackage{hhline}
\usepackage[utf8x]{inputenc}
\usepackage{longtable}
\usepackage{multirow}
\usepackage{rotating}
\usepackage{setspace}
\usepackage{subcaption}
\usepackage{tabularx}
\usepackage{url}
\usepackage{tablefootnote}

\newcolumntype{Y}{>{\RaggedRight\arraybackslash}X}

\usepackage{array}
\newcolumntype{M}[1]{>{\centering\arraybackslash\hspace{0pt}}m{#1}}

\begin{document}

\begin{frontmatter}

\title{Deep convolutional neural networks for brain image analysis on magnetic resonance imaging: a review}

\author[]{Jose Bernal\corref{mycorrespondingauthor}}
\cortext[mycorrespondingauthor]{Corresponding author}
\ead{jose.bernal@udg.edu}

\author[]{Kaisar Kushibar}
\ead{kaisar.kushibar@udg.edu}

\author[]{Daniel S. Asfaw}
\ead{daniel.asfaw@udg.edu}

\author[]{Sergi Valverde}
\ead{svalverde@eia.udg.edu}

\author[]{Arnau Oliver}
\ead{aoliver@eia.udg.edu}

\author[]{Robert Mart\'i}
\ead{marly@eia.udg.edu }

\author[]{Xavier Llad\'o}
\ead{llado@eia.udg.edu}

\address[mymainaddress]{Computer Vision and Robotics Institute\\
	Dept. of Computer Architecture and Technology\\
	University of Girona\\
	Ed. P-IV, Av. Lluis Santal\'o s/n, 17003 Girona (Spain)}

\begin{abstract}
In recent years, deep convolutional neural networks (CNNs) have shown record-shattering performance in a variety of computer vision problems, such as visual object recognition, detection and segmentation. These methods have also been utilised in medical image analysis domain for lesion segmentation, anatomical segmentation and classification. We present an extensive literature review of CNN techniques applied in brain magnetic resonance imaging (MRI) analysis, focusing on the architectures, pre-processing, data-preparation and post-processing strategies available in these works. The aim of this study is three-fold. Our primary goal is to report how different CNN architectures have evolved, discuss state-of-the-art strategies, condense their results obtained using public datasets and examine their pros and cons. Second, this paper is intended to be a detailed reference of the research activity in deep CNN for brain MRI analysis. Finally, we present a perspective on the future of CNNs in which we hint some of the research directions in subsequent years.
\end{abstract}


\end{frontmatter}


\section{Introduction}
\label{intro} 

Convolutional neural networks (CNNs), an outstanding branch of deep learning applications to visual purposes, have earned major attention in the last years due to its breakthrough performances in varied computer vision applications, such as in object recognition, detection and segmentation challenges~\cite{russakovsky2015imagenet, lin2014microsoft, everingham2015pascal}, in which they have achieved astonishing performances~\cite{krizhevsky2012imagenet, he2015deep, szegedy2016inception, noh2015learning, chen14semantic}. Their success has not been limited to reach top positions in different tasks, but also to achieve competent personnel levels in visual object recognition tasks~\cite{ioffe2015batch} and, perhaps more importantly, sensitive medical applications~\cite{esteva2017dermatologist}. 

\begin{figure}
    \centering
    \includegraphics[width=0.6\textwidth]{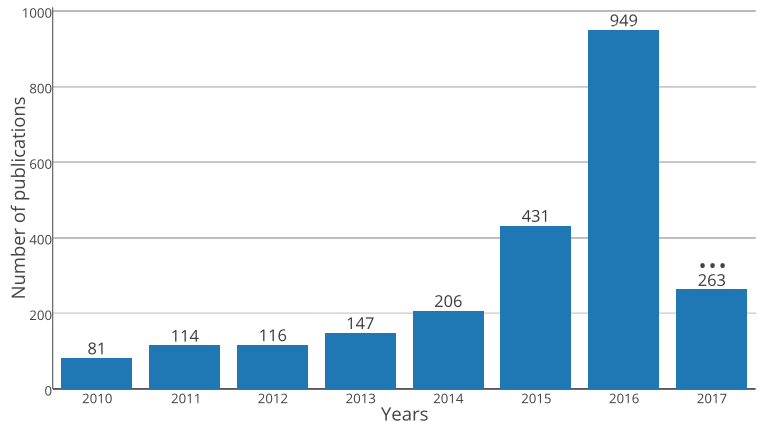}
    \caption{Number of publications per year in IEEE-Xplore containing “deep learning” and “medical imaging” keywords from 2010 to 2017. (Queried: June 6th, 2017). \label{fig:deep-learning-in-medical}}
\end{figure}

CNNs have been used in medical imaging applications since the 1990s in areas such as lung structure and nodule detection \cite{hasegawa1994convolution,lo1995artificial} and breast tissue classification \cite{sahiner1996classification}. However, due to the lack of labelled training data and computational power limitations by that time, it was not possible to train deep CNNs without over-fitting. As a result, proposals in this regard were discontinued in the field for some years. With the time, large annotated training datasets and more powerful graphics processing units (GPUs) have been created, enabling researchers to continue working in the area. This trend can be observed in Fig.~\ref{fig:deep-learning-in-medical} in which the number of papers using these strategies has increased year after year from 2010. In brain image analysis, the pioneering work appeared in multiple sclerosis (MS) by Maleki \etal~\cite{maleki2012diagnosis}, after Krizhevsky \etal~\cite{krizhevsky2012imagenet} rekindled research on CNNs in 2012. Nowadays, deep CNN architectures are widely used in brain MRI for preprocessing data~\cite{kleesiek2016deep}, detecting and segmenting lesions~\cite{dou2016automatic, roth2014new, sirinukunwattana2016locality, anthimopoulos2016lung, albarqouni2016aggnet, brosch2016deep} and segmenting tumours~\cite{pereira2016brain, kamnitsas2016efficient,havaei2016brain}, whole tissue~\cite{moeskops2016automatic, zhang2015deep, chen2016voxresnet} and sub-cortical structures~\cite{Dolz2017, Wachinger2017}.

Although in medical image classification tasks, there are usually fewer classes compared to large-scale semantic image recognition, two significant difficulties hindering achieving similar accuracy to human raters. First, there is a lack of sufficiently labelled training data. Essentially, generating highly accurate labels and finding sufficient pre-processed and representative data require a considerable amount of time. Second, medical image annotation is carried out by experts which is subjective and error-prone~\cite{llado2012automated,llado2012segmentation}. Learning a model from a less accurate representation of training samples degenerates the algorithm accuracy.

Brain MR image analysis has traditionally been an important area of research, attracting researchers to work on different tasks, such as lesion detection and segmentation, tissue segmentation and brain parcellation on neonatal, infant and adult subjects~\cite{gonzalez2016review,mitra2014lesion,roura2015toolbox,shen2012detecting}. Several public brain MR image datasets are available to the community, especially those organised by the Medical Image Computing and Computer-Assisted Intervention (MICCAI\footnote{http://www.miccai.org/}) society, actively encouraging research and publications in the field. These standard evaluation frameworks have been proposed for quantitatively comparing brain segmentation algorithms
under the same directives and conditions. Consequently, the number of publications using CNNs in brain MRI has been increasing: of 23 works collected from PubMed, Scholar, IEEE Xplore and Scopus databases for this review work, 20 of them were published in the period from 2015 to May 2017. 

Deep learning methods have been extensively reviewed in recent years~\cite{bengio2013deep, deng2014deep, schmidhuber2015deep, bengio2013representation, lecun2015deep} and discussed for particular applications~\cite{deng2016deep,guo2016deep,angermueller2016deep,mamoshina2016applications}.  Developments in the field of medical image analysis have been covered in~\cite{shen2017deep, litjens2017survey}. To the best of our knowledge, this work is the first to present a detailed review of deep CNN techniques applied to brain MRI analysis. We intend to comprise all publications in this field published from 2012 to June 2017. Along with~\cite{shen2017deep, litjens2017survey}, we aim to provide a good reading basis for newcomers to the topic. 

The rest of this review is structured as follows. The general concepts in CNNs are given in Section~\ref{sec:revision}. Note that an experienced reader could skip this section. Afterwards, the different methods published in the literature on brain MRI are surveyed and analysed regarding advantages and disadvantages in Section~\ref{section3}. In Section~\ref{section5}, evaluations and comparisons of the works are performed, based on the reported numerical results. We conclude this work with a discussion, indicating future trends in the field.

\section{Deep convolutional neural networks \label{sec:revision}}
For years, conventional supervised machine-learning techniques were built using automatic learning techniques and well-engineered algorithms. The approach consisted in taking the raw data, describing its content with low-dimensional feature vectors -- using specific prior knowledge of the addressed scenario -- and inputting the vectors to a trainable classifier. While the classifier was indeed useful for other purposes, the features were not necessarily generic. Indeed, the overall
accuracy of the method would depend on how appropriately designed were the heuristics~\cite{lecun1998gradient}. 

Representation learning appears as an alternative to this drawback: discover automatically
suitable detection and classification representations from the input data. One of the first successful attempts using this strategy took place in 1998 when LeCun \etal~\cite{lecun1990handwritten} presented a five-layer fully-adaptive architecture for addressing handwritten digit recognition. Despite its accuracy
results (1\% error rate and 9\% reject rate from a dataset of around 1000 samples), the authors were able to apply neural networks to a real-world task. However, these authors were not able to scale a large number of hidden layers and larger images, mainly due to computational resource constraints and the ``vanishing gradient'' problem. This latter situation being the case when the gradients of the network's output with regard to the parameters in early layers becomes negligible \cite{279181}. 

Two initial solutions were proposed in the literature to address the ``vanishing gradient" problem. The first one was introduced by Hinton \etal~\cite{hinton2006fast}, in which training was conducted in two stages. The first stage was training each layer in an unsupervised, greedy manner, essentially initialising the parameters with better values than random or uniform values. In the second stage, the network was fine-tuned using labelled data. This work rekindled interest in current deep learning strategies. The second solution appeared with the introduction of the ReLU activation function~\cite{glorot2011deep}, which showed successful results in large-scale image recognition tasks~\cite{krizhevsky2012imagenet}. Along with the availability of faster GPUs, these strategies have dramatically increased the research in deep neural networks.

One of the most widely adopted approaches of deep neural networks is the convolutional neural networks which can process array-like data~\cite{lecun2015deep}, such as images or video sequences. From a high-level perspective, the idea behind CNN is to identify compositional hierarchy features which object from the real world exhibit: low-level features (e.g. edges) form patterns, and these specific patterns form more intrinsic structures (e.g. shapes, textures). Further information about the building blocks of CNN is provided in following sections.

\begin{figure}[t]
    \centering
    \includegraphics[width=\textwidth]{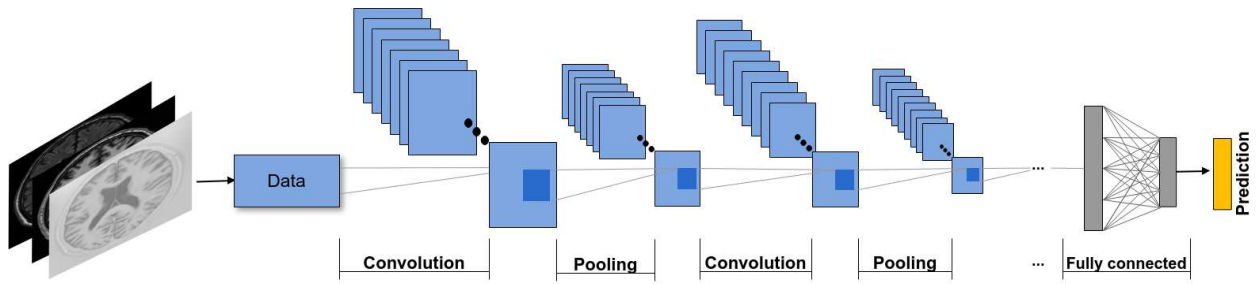}
    \caption[]{Generic architecture of convolutional neural networks. The output of each convolution operation at each layer is activated using activation functions before applying pooling operations. The convolution operation produces different numbers of feature maps, depending on the numbers of filters used. The pooling operations reduce the spatial dimensions of each feature map. After convolution and pooling the layers, the future maps are flattened in the fully connected layer before a prediction is made using linear classifiers.} 
    \label{fig:2DCNN}
\end{figure}

\subsection{Building blocks of CNN}
\subsubsection{Convolutional layer}
CNNs learn the relationships among the pixels of input images by extracting representative features using convolution and pooling operations. The features detected at each layer using learnt kernels vary concerning their complexity, with the first layer extracting simple features, such as edges, and the later layers extracting more complex and high-level features. The convolution operation in CNNs has three main advantages. First, the weight-sharing mechanism helps to deal with high dimensional data, either 2D images or 3D data, such as videos and volumetric images. Second, local connectivity of the input topology can be exploited using 2D or 3D kernels. Finally, slight shift invariance is achieved using pooling layer. The typical architecture of a 2D CNN is shown in Fig.~\ref{fig:2DCNN}.

Very deep CNN architectures were recently proposed to replace the conventional convolutional layer with modules with more powerful representation while using less computational resources. For instance, Szegedy \etal~\cite{szegedy2015going} introduced inception modules that could extract multi-scale features from the input feature maps and could efficiently reduce the number of parameters. This module was further improved in \cite{ioffe2015batch} and \cite{szegedy2015rethinking}. The recent version of the inception module in \cite{szegedy2016inception} was created to have a more uniform simplified architecture than previous versions and thus could achieve top performance in large-scale image classification tasks.

\subsubsection{Non-linearity layer}
The above convolutional layer is usually followed by non-linearity operations. Non-linearity is achieved using a specific family of functions called activation functions. These activation functions ensure that the representation in the input space is mapped to a sparse one and, hence achieving (i) a certain invariance to data variability and (ii) a computationally efficient representation~\cite{glorot2011deep}. The former situation refers to the fact that sparse representations are more resilient to slight modifications than dense ones. In the past, sigmoid and hyperbolic tangent functions were commonly used for this purpose. However, for large-scale image recognition, novel activation functions are being continuously proposed. We categorise the commonly used activation functions into three broad families.

\textbf{Rectified linear units (ReLUs) and variants:} 
which are expressed in a general form as
\begin{equation}
f\left(z_{ lk}^{xy}\right) = \begin{cases}
\max\left( z_{lk}^{xy},\, 0\right)\qquad\quad \text{if } z_{ lk}^{xy} > 0,\\
\min\left(\alpha \cdot z_{lk}^{xy},\, 0\right)\qquad \text{if } z_{ lk}^{xy} \leq 0,\\
\end{cases}
\end{equation}
where $z_{ lk}^{xy}$ is the input value at position $ (x,y)$ on the $ k^{th}$ feature map at the $ l^{th }$ layer and $\alpha$ is the slope of the negative linear function. There are five special cases distinguished depending on the conditions over $\alpha$. First, if $\alpha = 0$, the expression results in the so-called ReLU~\cite{krizhevsky2012imagenet, nair2010rectified} which is one of the most commonly used activation functions~\cite{lecun2015deep}. Despite its computationally efficient implementation, this method presents some drawbacks due to its gradient discontinuity at the origin in terms of gradient update and empirical performance~\cite{szegedy2016inception, he2015delving}. Second, if $\alpha$ is a small constant, the variant is referred as Leaky ReLU (LReLU)~\cite{maas2013rectifier}. This approximation enables to cope with the problem of zero gradient.
Third, if $\alpha$ is tuned up in the training process along with other parameters using back-propagation, the approach is referred as 
Parametric ReLU (PReLU)~\cite{he2015delving}. Fourth, in~\cite{xu2015empirical}, the parameter $\alpha$ is sampled from a uniform distribution for each example, and this approach is called Randomised ReLU (RReLU). Although using a slope parameter for the negative part in ReLU showed improvements in performance, RReLU showed better performance than the other ReLU variants, based on the evaluation in~\cite{xu2015empirical} on image classification tasks. Fifth, recently, Jin  \etal~\cite{jin2016deep} proposed a new type of activation function, called S-shaped ReLU (SReLU), in which the essential idea is to consider a piecewise function composed of three linear functions, i.e.
\begin{equation}
f\left(z_{ lk}^{xy}\right) = \begin{cases}
t_r + a_r \cdot \left( z_{lk}^{xy} - t_r\right)\qquad \text{if } z_{ lk}^{xy} \geq t_r,\\
z_{lk}^{xy}\qquad\qquad\qquad\qquad \text{if } t_r > z_{ lk}^{xy} > t_l,\\
t_l + a_l \cdot \left( z_{lk}^{xy} - t_l\right)\qquad\,\, \text{if } z_{ lk}^{xy} \leq t_l,\\
\end{cases}
\end{equation}
where $a_r$, $a_l$, $t_r$ and $t_l$ are learnable parameters. According to the experiments in the paper, the SReLU is able to learn both convex and non-convex functions. This activation improved the performance of well-known CNN architectures in MNIST~\cite{lecun1998mnist} and ImageNet~\cite{deng2009imagenet} datasets, compared to ReLU, LReLU and PReLU.
 
\textbf{Maxout and variants:} Maxout~\cite{goodfellow2013maxout} was proposed in particular to improve the optimisation and model averaging performance of dropout training. This activation function is also a generalisation of ReLU. It is computed by calculating the maximum across K affine feature maps, i.e.

\begin{equation}
 \label{Maxout}
f\left(z_{ lk}^{xy}\right) = \max_{k\in[1, K]}\left(z_{ lk}^{xy}\right).
\end{equation}

One major drawback of this technique is that the number of weights to be learned in each layer is increased by a factor of $K$. A workaround to this situation was proposed by Springenberg and Reidmiller \cite{springenberg2013improving} in which a probabilistic sampling procedure to compute the maximum across feature maps, generalising maxout was considered. This activation function called Probout empirically matched or improved the performance of maxout~\cite{springenberg2013improving}. 

\textbf{Exponential Linear Units (ELU) and variants:} ELU\cite{chen2015automatic} are similar to the extensions of ReLU as they employ an identity for positive inputs. Unlike ReLU variants, they provide saturated output for negative inputs. The saturation in the negative regions of the function is reported to be beneficial for expediting the learning and improving the performance of very deep CNNs. It is defined as

\begin{equation}  
 \label{Probout} 
f\left(z_{ lk}^{xy}\right) = \max\left(z_{ lk}^{xy} ,\,0\right) + \alpha\cdot\min\left(e^{z_{ lk}^{xy}}-1,\,0\right).
\end{equation} 

Trottier \etal~\cite{trottier2016parametric} defined parameters controlling different aspects of the ELU function and proposed learning them with gradient descent during training. This parametric ELU (PELU) further improved the speed and performance of training deep networks. Using off-the-shelf ResNet~\cite{he2015deep}, PELU performed better than ELU and ReLU in image classification tasks on MNIST, CIFAR-10/100 and ImageNet datasets \cite{trottier2016parametric}.

\subsubsection{Pooling and sub-sampling layer}
Typical convolutional layers consist of three steps. First, the layer performs several convolutions to produce feature maps. Second, non-linear activation functions are used on the resulting maps. 
Third, the output is modified by the pooling layer before reaching the next convolutional layer. The idea of a pooling function is to extract a summary statistics of non-overlapping neighbourhoods -- usually -- to (i) reduce the number of parameters in the following layers, (ii)
control over-fitting, and (iii) achieve slight translation  invariance~\cite{Goodfellow-et-al-2016-Book}.

Among several pooling methods, max pooling and average pooling are widely used types. Their operations are similar except that the former uses the maximum of the activations, and the latter uses the average of them. Max pooling is the most common due to its empirical performance \cite{boureau2010learning, boureau2010theoretical}. Apart from these two approaches, other pooling methods are proposed to achieve higher performance.

\textbf{Stochastic Pooling:} Zeiler \etal~\cite{14d852e0427f418c8cba34d967623a3e} proposed stochastic pooling to regularise the convolutional layers and, hence, overcome the overfitting problem of average pooling and max pooling. In this method, the output of the activation from each pooling region is selected first by computing the probability distribution $p$ by normalising the activations; then, the output activation is chosen by sampling from the multinomial distribution based on $p$. The authors showed that this pooling strategy improved the accuracy on MNIST, CIFAR-10/100 and street view house numbers image recognition (SVHN)~\cite{netzer2011reading} datasets. 

\textbf{Spatial pyramid pooling:} Another pooling method called spatial pyramid pooling~\cite{he2014spatial}, is proposed to work with any input image size. It is added after the last pooling layer, immediately before the fully connected layer, to generate fixed length representation regardless of the image size or scale. By simple modification of the off-the-shelf CNN architectures with a spatial pyramid pooling layer, this pooling scheme was reported to improve the performance of CNN models.

\textbf{Deformation pooling:} Despite max pooling and average pooling being useful in handling deformations, they are incapable of learning deformation constraints or geometric models from object parts. Deformation pooling (referred as def-pooling) was explicitly designed to overcome this drawback for object detection applications~\cite{ouyang2015deepid}.

\textbf{Combination of max and average pooling:} Lee \etal~\cite{lee2016generalizing} proposed a generalisation of max and average pooling operations that allowed them to be combined and adjusted in the training process. The combination operation was performed using mixed max-average and gated max-average pooling. The latter approach improved the characteristics of the region being pooled. The learning is performed during a combined pooling operation using a binary tree in which each leaf of the tree is associated with a learnt pooling filter. This strategy is called tree pooling. An experiment in~\cite{lee2016generalizing} reported that this method boosted the performance of AlexNet \cite{krizhevsky2012imagenet} and GoogLeNet \cite{szegedy2015going} models on MNIST, CIFAR-10 and SVHN datasets. 

\subsubsection{Fully connected layer \label{Fullycoon}}
Unlike convolutional layer, the Fully Connected (FC) layer has a full connection to all of the units in the previous layer, as shown in Figure~\ref{fig:2DCNN}. This layer changes the previous layers 2D structure features into a predefined one-dimensional feature vector. Essentially, the main task of the FC layer is to mine the incoming features to extract information about the content of the input image. The process usually consists in
flattening the feature maps coming from convolutional layers, to achieve a one dimension feature vector representation, and, then, inputting it into the FC layer. The output of this layer could
either be the predict class labels~\cite{krizhevsky2012imagenet} or an intermediate layer~\cite{girshick2014rich} (consecutive FC layers can be stacked together). 

Implementing FC layers usually require a large number of parameters -- compared to other layers -- as each cell of a feature map is fully connected to all elements in the previous layer. Besides, there are two drawbacks of these kinds of layers: (i) a single output is produced – if it is used as output layer – and (ii) accepts fixed-size inputs. The former issue means that a single input image receives a single output label. This situation is computationally inconvenient if the CNN is intended to be used in segmentation tasks rather than classification ones. The second issue relates to problems on extending the network: either the input images are scaled to fit the requirements of the network or the network is re-factorised to be able to process the new data. A solution to these problems lies in the fact that FC layers can be converted to convolutional layers of $1\times 1$ kernels~\cite{long2015fully} ($1\times 1\times 1$ kernels in case of 3D). In this way, the model keeps the fully connected functionality while accepting arbitrary input size image and making dense predictions. These types of architectures, called fully convolutional networks (FCNN), have been gradually introduced in the literature~\cite{dou2016automatic,brosch2016deep, kamnitsas2016efficient, havaei2016brain, Dolz2017,long2015fully}.

\subsubsection{Loss function \label{error_fun}}
CNNs are well-known for their ability to extract discriminative features using learned weights in each layer. The learning process is reinforced by employing appropriate loss functions. Loss functions are designed to encourage intra-class similarity and inter-class separability.

In image classification tasks, most CNNs employ softmax loss, which is a combination of the softmax function and cross-entropy loss, mainly because of simplicity and the probabilistic interpretation of softmax classifiers. Hinge loss is another type of loss function, which in conjunction with margin-based classifiers was reported to perform better than cross-entropy loss in a standard image classification challenge~\cite{tang2013deep}. More recently, the research in~\cite{liu2016large} proposed a Large-Margin Softmax Loss (L-softmax), which is a modification of the softmax loss with a distance margin constraint. This loss was reported to boost the performance of deeply learned features in visual classification and verification tasks.

One of the main issues during the training phase is the disproportion among class observations (i.e. the number of available samples per class varies dramatically) as the resulting classifier may be biased towards the majority class. For instance, the region of interest to be analysed occupies only a small part of the scan. This problem can be addressed in the training phase by (i) undersampling the majority class~\cite{kamnitsas2016efficient,moeskops2016automatic}, (ii) merging or subdividing classes~\cite{chen2016voxresnet} and (iii) penalising sample misclassification based on the reciprocal frequency~\cite{cruz2016tackling}; or throughout the loss function. Brosch \etal~\cite{brosch2015deep} considered an approach for brain lesion segmentation in which the weighted sum of the mean squared difference of the lesion voxels (sensitivity) and non-lesion voxels (specificity) was used. Also, Milletari \etal~\cite{milletari2016v} proposed an objective function based on Dice Similarity Coefficient (DSC) for a two-class problem. According to the authors, this loss function yielded higher performance than the same architecture using softmax.

Other loss functions are task-specific, such as contrastive losses, which are mostly used to measure similarities between two data points. Su \etal~\cite{sun2014deep} used a combination of softmax and contrastive losses for face recognition. The purpose of the softmax loss was to classify faces into different classes, thus encouraging inter-class separation. Then, the contrastive losses were applied for face verification to enforce intra-class similarity. A distance constraint was introduced to the contrastive loss in~\cite{schroff2015facenet} to obtain more discrimination of features.

\subsubsection{Regularisations}
\label{regularization} 
We observed the appealing performance of deep CNN methods in different domains, although they use enormous numbers of parameters. Unless trained on a large, labelled training dataset, proper regularisation should be employed to mitigate over-fitting. There are several regularisation methods widely used in the community, such as $L_1$ or $L_2$ regularisation approaches encouraging sparsity and small weight magnitude; early stopping~\cite{Goodfellow-et-al-2016-Book} forcing the training to stop when there is a sign of over-fitting (also widely used to select hyper-parameters of the model); batch normalisation~\cite{ioffe2015batch} in which each batch is preprocessed to achieve mean equal to zero and standard deviation equal to one; and dropout~\cite{wager2013dropout,srivastava14a} in which some feature map units are skipped. This last approach being the dominant as (i) it is computationally inexpensive and (ii) prevents co-adaptation among feature map units by encouraging independent contributions of each of the units to the final prediction.

Different improvements in dropout have been proposed. For instance, Wang \etal~ \cite{wang2013fast} proposed a method to expedite training using dropout by sampling from a Gaussian approximation, instead of repeatedly sampling a random subset of input features. Another research in~\cite{wan2013regularization} proposed Dropconnect, which drops a subset of weights within the network instead of subset of activation within each layer. This approach showed better generalisation than the original dropout in standard image recognition tasks.

Data augmentation is another approach to improve the generalisation of CNNs by increasing the training dataset by artificially generating data. Although different image transformations~\cite{krizhevsky2012imagenet,szegedy2015going,he2015delving,howard2013some} and colour perturbation \cite{chatfield2014return} have been used and proved useful in recognition tasks, one limitation with these methods is that no theoretical background indicates the group of transformations helping to improve the generalisation capability of a model.

\subsubsection{Optimisation}
Deep CNNs are learnt by searching appropriate values for model parameters optimising the loss function. Gradient descent methods, in which the parameter update is performed using a back-propagation algorithm, are widely used for minimisation~\cite{lecun1989backpropagation}. This parameter update is undertaken by computing the loss function for a single, small subset, or the whole training set. Each of these cases is referred in the literature as stochastic, mini-batch and batch gradient descent, respectively. Updating values using the whole training set can be computationally expensive and, hence, mini-batch gradient descent is ubiquitously used in the community, leading to smoother parameter updating and more stable convergence.

The primary difficulty when optimising deep CNNs using gradient descent-based algorithms is the non-convex nature of the loss function. Non-convex functions have several local minima, in which gradient descent methods could get easily trapped~\cite{sutton1986two}. Dauphin \etal~\cite{pascanu2014saddle} argued that saddle points, where the gradient vanishes at non-local optimum places, are much more of an issue than local optima in optimising non-convex loss functions. The momentum variable was introduced to the stochastic gradient descent (SGD) update method to avoid oscillation in local optima. Further, momentum-based SGD was improved to have some prescience about the next update direction, referred to as Nesterov accelerated gradient descent~\cite{nesterov1983method, sutskever2013importance}. Recent works have also been proposed to escape saddle points during the optimisation process~\cite{ge2015escaping, anandkumar2016efficient}.

One major drawback of the momentum-based SGD methods is to select an adequate learning rate, a parameter determining how substantial a change in the update should be made. This parameter is commonly set globally to be equal for all settings. Much work has been carried out on tuning the global learning rate adaptively based on the gradient of each parameter. One method in~\cite{duchi2011adaptive}, called adaptive gradient algorithm (Adagrad), scales the learning rate of each parameter according to the sum of the previous gradients. The amount of updating differs based on the sparsity of the parameters in each gradient update. Adadelta~\cite{zeiler2012adadelta} and Root Mean Square Propagation (RMSProp)~\cite{tieleman2012lecture} improved the drawbacks of Adagrad unstable gradient updating by adopting a more stable learning rate scheme. Kingma and Ba \cite{kingma2014adam} proposed Adam, which improved the gradient computation of both Adadelta and RMSProp. They also empirically showed improved performance in non-convex optimisation problems in different machine learning tasks.

\subsubsection{Weight initialisation}
Weight initialisation is crucial for non-convex optimisation algorithms. Initialising all the parameters to the same small value results in the undesired case in which all weights are updated the same during back-propagation. Another option that could break the symmetry is to initialise the weights with small values taken from a Gaussian distribution~\cite{krizhevsky2012imagenet}. Glorot and Bengio~
\cite{glorot2010understanding} proposed a method known as ``Xavier" initialisation that normalises the variance of each neuron's output to one. There are two beneficial outcomes from this approach: (i) this avoids the variance in each layer and (ii) keeps the weights from exploding to large values or vanishing. This approach was later improved in~\cite{he2015delving} to account for the non-linearity of ReLU. This allows training very deep networks with better convergence than the ``Xavier" initialisation. Another recent work from He \etal~\cite{he2015delving} proposed an initialisation by explicitly modelling the non-linearity of rectifiers (ReLU/PReLU), which helped the convergence of extremely deep models.

\subsubsection{Normalisation}
It is a common practice to use mean-centred training datasets by applying contrast normalisation to train deep CNNs~\cite{krizhevsky2012imagenet,pereira2016brain,kamnitsas2016efficient,havaei2016brain,lin2013network}. This simple preprocessing technique improves the convergence speed of SGD algorithms. Ioffe \etal~\cite{ioffe2015batch} argued that training deep networks can be slow since the distribution of parameters across hidden units changes dynamically during training, which is a phenomenon called the internal covariate shift. Their proposal consisted in normalising the data input in each layer, with a technique known as batch normalisation. This method provides any layer in a network with inputs that come from a unit Gaussian distribution and also enables the use of a high learning rate with a less careful weight initialisation choice. More recently, in~\cite{salimans2016weight}, the authors introduced weight normalisation. With simple weight re-parametrising at each layer, they were able to show improved conditioning of the optimisation problem and faster convergence of SGD methods.

\section{CNN methods for brain image analysis \label{section3}}
Automatic segmentation in brain MR has been widely investigated throughout the years to support medical tasks, such as diagnosis and patient monitoring. In the literature, most of the deep learning-based papers for brain MRI analysis have focused on lesion, tumour, tissue and whole brain and sub-cortical structure segmentation. This trend is primarily attributed to the different MICCAI challenges. Each year, the number of participants in these challenges employing deep learning approaches has been increasing. In this section, the proposed methods in this domain are discussed in more detail, based on their CNN architectures. Additionally, the description comprises the considered pipeline steps: pre-processing, data preparation, classification and post-processing techniques. There is no specific CNN architecture that is only suitable for a particular application; rather, a CNN model that is proposed for tumour segmentation could work for structure segmentation, and vice versa, with little or no modification. 

\begin{figure*}[t]
    \centering
    \includegraphics[width=.99\textwidth]{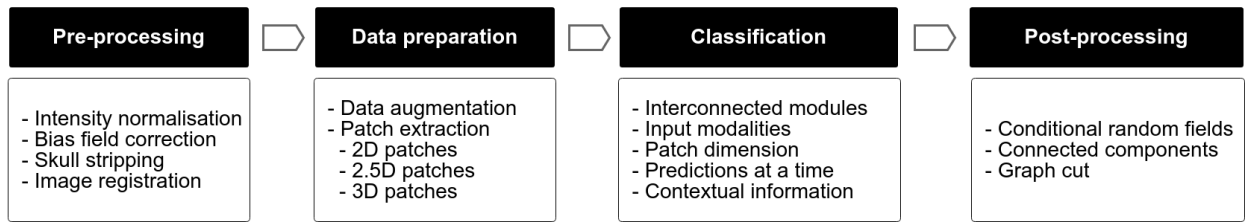}
    \caption[]{A general block diagram of the CNN-based image analysis pipeline.\label{fig:general-pipeline}}
\end{figure*}

The segmentation methods proposed in the literature could be seen from a top-level perspective as presented in Fig.~\ref{fig:general-pipeline}. The overall pipeline consists of four stages: preprocessing, data preparation, classification, and post-processing. In the preprocessing stage the different pipelines consider noise filtering techniques, inter and intra-patient normalisation and skull-stripping methods -- when necessary. Then, the data is prepared to be processed by the classifier. For instance, data preparation could contemplate augmenting the data or, in patch-based strategies, extracting patches from the input volumes. After that, classification takes place. Finally, once the segmentation results are obtained, they could be refined by removing small isolated areas by selecting the biggest groups only or smoothing regions. It is important to note that some works do not specify any preprocessing or post-processing methods. In the following sections, we will discuss each of these blocks emphasising on the CNN strategies.

\subsection{Pre-processing}
Pre-processing in MRI is an essential step for subsequent segmentation task. In training supervised models such as CNNs, the input training data hugely influences the performance of the model, so having preprocessed and well-annotated data is a crucial step in achieving good performance. 

\begin{figure}
    \centering
    \begin{subfigure}[b]{0.22\textwidth}
        \includegraphics[width=\textwidth]{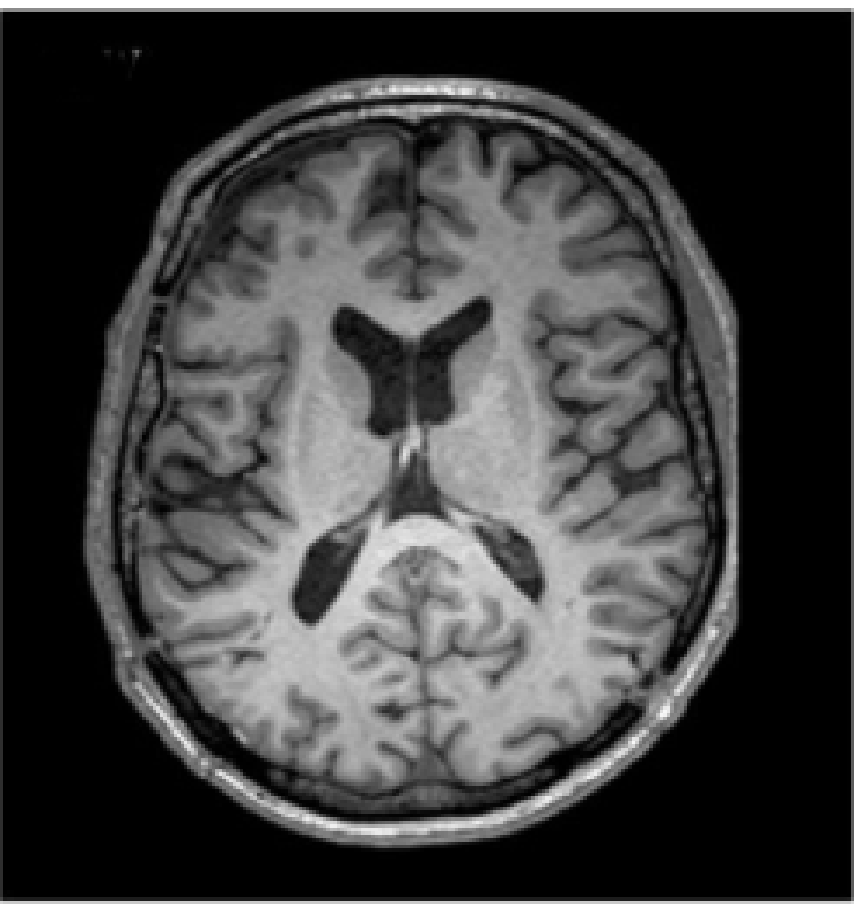}
        \caption{T1-w image\label{fig:original-volume}}
    \end{subfigure}
    ~
    \begin{subfigure}[b]{0.22\textwidth}
        \includegraphics[width=\textwidth]{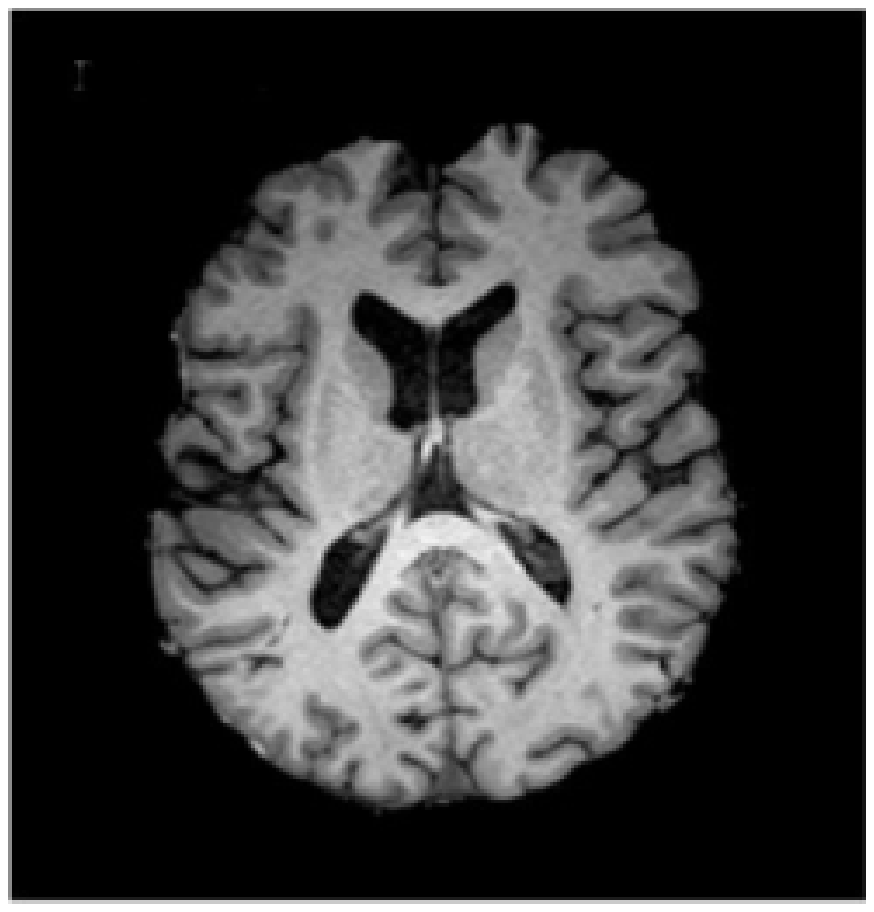}
        \caption{Skull stripped\label{fig:skull-stripping}}
    \end{subfigure}
    ~
    \begin{subfigure}[b]{0.22\textwidth}
        \includegraphics[width=\textwidth]{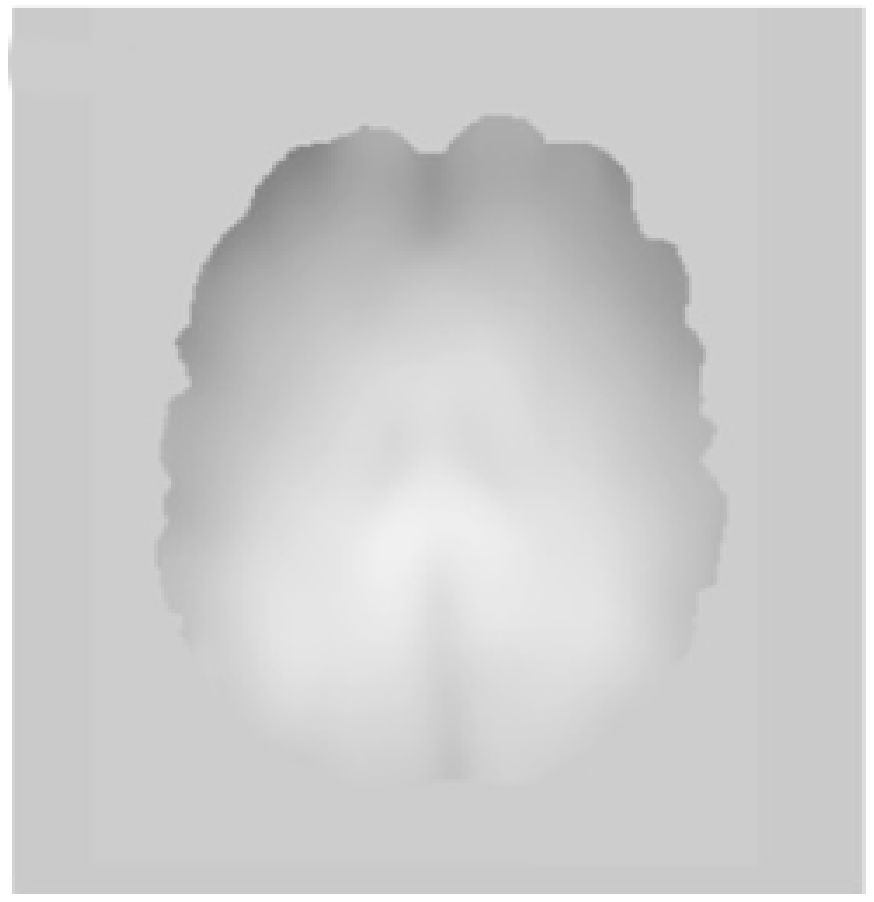}
        \caption{Bias field\label{fig:bias-field-correction}}
    \end{subfigure}
    ~
    \begin{subfigure}[b]{0.22\textwidth}
        \includegraphics[width=\textwidth]{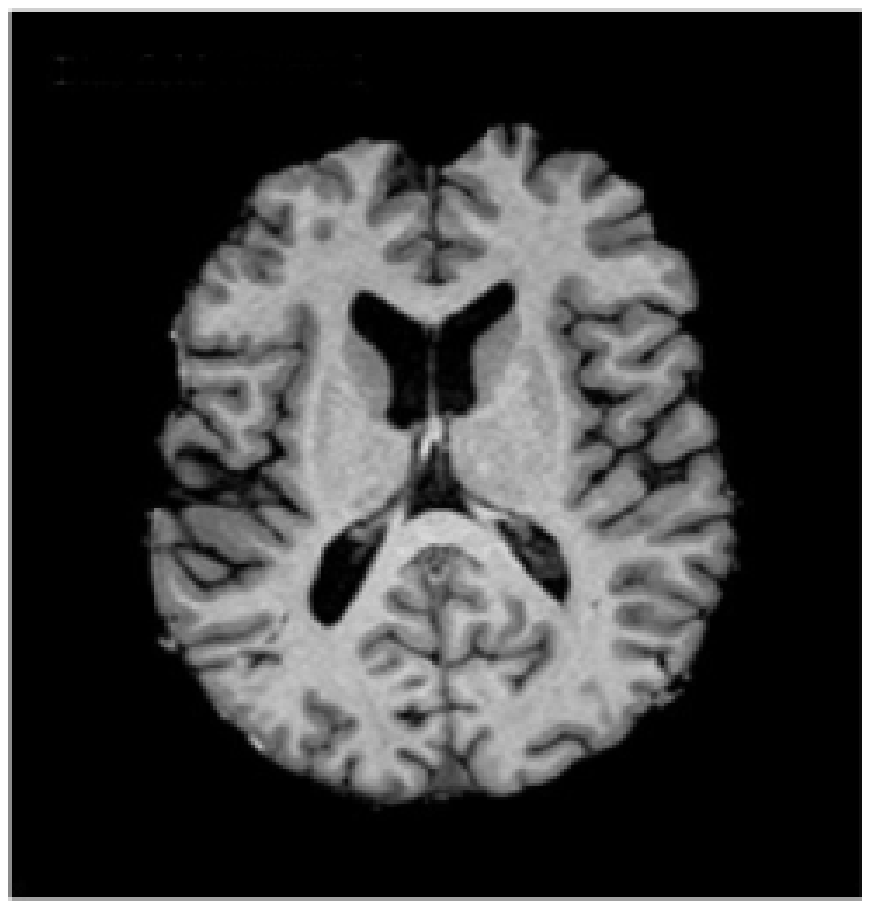}
        \caption{Preprocessed T1-w\label{fig:preprocessed}}
    \end{subfigure}
    \caption[An example of preprocessing methods]{An example of two preprocessing methods: skull stripping and bias field correction~\cite{despotovic2015mri}. In the figure, a T1-w slice is displayed in (a), brain tissue after removing non-brain areas in (b), estimated bias field in (c), and preprocessed brain tissue in (d). \label{fig:preprocessing}}
\end{figure}

Acquired brain MRI volumes incorporate non-brain tissue parts of the head, such as eyes, fat, spinal cord or skull. The process of extracting the brain tissue from non-brain one is referred in the literature as skull stripping. An example of an original volume and its corresponding skull stripped output is presented in Fig.~\ref{fig:original-volume} and ~\ref{fig:skull-stripping}, respectively. This step has direct consequences on the performance of automated methods, as the inclusion of skull or eyes as brain tissue may lead to unexpected results in classification~\cite{acosta2008impact, popescu2012optimizing}, while unintended removal of the cortical surface may result in underestimation of the cortical thickness~\cite{sadananthan2010skull}. Among the different methods proposed in the literature for skull-stripping~\cite{lee2003evaluation, acosta2008impact, roura2014marga}, methods such as BET~\cite{BET2002,BET22005}, BSE~\cite{shattuck2001magnetic}, ROBEX~\cite{iglesias2011robust} and BEaST~\cite{Eskildsen20122362} are commonly used. In the literature, the methods used in clinical trail datasets employed BET~\cite{brosch2016deep,moeskops2016automatic, yoo2014deep} and ROBEX~\cite{kamnitsas2016efficient}. Zhang \etal~\cite{zhang2015deep} applied a paediatric brain skull stripping algorithm known as LABEL~\cite{shi2012label}. The public dataset images from the Brain Tumor Image Segmentation Challenge (BRATS)  2013\footnote{\url{http://martinos.org/qtim/miccai2013/}}, 2014\footnote{\url{https://sites.google.com/site/miccaibrats2014/}} and 2015\footnote{\url{https://www.smir.ch/BRATS/Start2015}} are preprocessed in this regard beforehand.

Inherent characteristics of the MRI acquisition process such as differences in the magnetic field, bandwidth filtering of the data or eddy currents driven by field gradients usually result in image artefacts that may also have a negative impact on the performance of the methods~\cite{simmons1994sources}. There is the need to remove spurious intensity variations caused by inhomogeneity of the magnetic fields and coils. In these cases, intensity correction of the MRI images is performed either before tissue segmentation, or as an integrated part of the tissue segmentation pipeline. A common technique to address this problem is to use bias-field correction~\cite{juntu2005bias}. The estimated bias field and the corrected version of Fig.~\ref{fig:skull-stripping} are depicted in Fig.~\ref{fig:bias-field-correction} and~\ref{fig:preprocessed}, respectively. Among the available strategies~\cite{arnold2001qualitative, ge2002age}, the non-parametric non-uniform intensity normalisation (N3)~\cite{sled1998nonparametric} and N4ITK~\cite{tustison2010n4itk} methods are currently the most widely used. Zhang \etal~\cite{zhang2015deep} and Yoo \etal~\cite{yoo2014deep} employed N3 algorithms on their clinical dataset. Similarly, Pereira \etal~\cite{pereira2016brain} used them in both BRATS 2013 and 2015 Challenges, Lyksborg \etal~\cite{lyksborg2015ensemble} in BRATS 2014, and Zikic \etal~\cite{zikic2014segmentation} in BRATS 2013. 

Brain MRI datasets might have volumes acquired from different scanner vendors and also from the same scanner but with different protocols. As a result, the volumes may exhibit non-uniform intensity representation for the same tissue types, i.e. interclass variability. To correct this problem, image normalisation algorithms are utilised. According to the literature, this intensity normalisation can be driven in two ways: (i) histogram matching~\cite{birenbaum2016longitudinal, urban2014multi, vaidhya2015multi,kleesiek2016deep,pereira2016brain} and (ii) normalise data to achieve zero mean and unit variance~\cite{pereira2016brain,havaei2016brain,chen2016voxresnet,Dolz2017,valverde2017improving}. In the former case, Urban \etal~\cite{urban2014multi} and Kleesiek \etal~\cite{kleesiek2016deep} considered matching the histogram of all volumes to a subject in the training set, which may result in fused grey levels, while Pereira \etal~\cite{pereira2016brain} -- based on the normalisation method proposed by Nyul \etal~\cite{nyul2000new} -- considered mapping to a virtual grey scale learnt directly from the data, so the undesired fusion of grey levels is avoided. Naturally, both normalisation strategies can be used one after the other one to improve the segmentation results. According to the results reported by Pereira \etal~\cite{pereira2016brain}, the preprocessing step improved their result, obtaining a mean gain of 4.6\%.

In addition to the above discussed pre-processing methods, image registration between different MRI modalities is important depending on the dataset analysed.  Image registration transforms different modalities of MRI into a common coordinate space. The authors of~\cite{maleki2012diagnosis,brosch2016deep,kamnitsas2016efficient, brosch2015deep}\footnote{The authors in~\cite{maleki2012diagnosis} and~\cite{brosch2015deep} reported using registration during their pre-processing stage but details were scarce in this regard.} applied image registration algorithms on their clinic trial dataset. For instance, Brosch \etal~\cite{brosch2016deep} applied a six degree-of-freedom intra-subject registration using one of the 3~mm scans as the target image to align the different modalities. Additionally, Kamnitsas \etal~\cite{kamnitsas2016efficient} applied affine atlas-based registration. 

\subsection{Data preparation}
\label{data_pre}
By data preparation, we refer to all of the operations performed before feeding the data into the network, such as data augmentation and patch extraction. Although this stage in most pipelines is considered to be pre-processing, we include it as a separate step since (i) the preprocessing steps are generic while these are particular for the CNN approaches and also (i) to provide more details.

Data augmentation is mainly employed to increase the training samples to mitigate over-fitting as discussed in Section~\ref{regularization}. It is a common practice to use data augmentation in computer vision tasks in which (i) the CNN architectures are very deep, and (ii) obtaining enormous amounts of labelled training data is difficult. Moreover, 
unless the dataset is large enough to correctly train the network (which is not the common case~\cite{havaei2016brain,lyksborg2015ensemble, vaidhya2015multi}), the high disproportion between the cardinality of the majority and the minority classes may lead to biased classifiers. Data augmentation could be used in these cases to alleviate and improve the overall performance of the model. Pereira \etal~\cite{pereira2016brain} used data augmentation for dual purposes: (i) to increase the training data by applying rotation at different angles and (ii) to introduce class balance by adding more data from the minority class. According to their experiments, by augmenting using rotation, they achieved better delineation of the complete tumour types, as well as of the intra-tumoural structures, reporting a mean gain of up to 2.6\%. In addition, by oversampling the minority class, they stated a mean gain of 1.9\%. In contrast, Havaei \etal~\cite{havaei2016brain} employed data augmentation by flipping the training data but had no success in improving the accuracy of the model. Furthermore, Chen \etal~\cite{chen2016voxresnet} used data augmentation as input, in addition to the given multi-modal images. The augmentation was carried out by subtracting a Gaussian smoothed version and applying histogram equalisation using the Contrast-Limited Adaptive Histogram Equalisation (CLAHE)~\cite{pizer1987adaptive} for enhancing local contrast~\cite{chen2016voxresnet, stollenga2015parallel}. Afterwards, they used the generated and original volumes to train their network. Nevertheless, the actual effect of this type of augmentation was not reported in their work.

The patch extraction can be performed on a single plane (referred as 2D architectures), from three anatomical planes (referred as 2.5D architectures) or directly from 3D. Also, the patches can be acquired from different imaging modalities -- if available. The advantages of one approach or the other one are discussed in the following section. After extracting the patches of training images, the data is normalised by subtracting the mean intensity and dividing by the standard deviation~\cite{brosch2016deep,pereira2016brain, zhang2015deep,chen2015automatic,yoo2014deep,dvorak2015structured}. The resulting zero mean and unit variance training data helps to expedite the convergence of SGD. It is also a recommended practice to shuffle the samples randomly, especially when training using a mini-batch gradient decent algorithm~\cite{bengio2012practical}.

\subsection{CNN architectures} \label{architectures}
The classification of CNN architectures in medical imaging can be grouped around five aspects: (i) number of interconnected operating modules, (ii) number of input modalities, (iii) input patch dimension, (iv) number of predictions at a time and (v) implicit and explicit contextual information. All these variants, their advantages and disadvantages are described in the following sections. A summary of the reviewed methods in the literature concerning pre-processing, post-processing and target of the segmentation are described in Table \ref{tab:architecture_summary}. 

\begin{table*}[t]
\centering
 \caption{Summary of deep learning approaches for MRI brain analysis with regard to the pre-processing, architecture, post-processing and target application. \label{tab:architecture_summary} }
    
     \resizebox{\textwidth}{!}
         { 
\begin{tabular}{l l l p{5cm} p{5cm} l}
\hline
 \multicolumn{2}{c}{\textbf{Architecture}} & \textbf{Article}             & \textbf{Pre-Processing} & \textbf{Post-processing} & \textbf{Target }  \\
\hline
\multirow{20}{*}{\rotatebox[origin=c]{90}{Single  CNN}} 
& 2D CNN & Maleki \etal~2012~\cite{maleki2012diagnosis} & Normalisation\newline Registration\newline Skull-stripping & - & MS lesions \\
& 2D CNN & Zikic \etal~2014~\cite{zikic2014segmentation} &  N4ITK  & -  & Tumour  \\
& 2D CNN & Dvorak \& Menze 2015~\cite{dvorak2015structured} & Normalisation\newline N4ITK & -  & Tumour  \\
& 2D CNN & Chen \etal~2015~\cite{chen2015automatic} &  Normalisation  & -  & Cerebral micro-bleed  \\
& 3D CNN & Lai 2015~\cite{lai2015deep} & Masking in normalised coordinates & Connected component analysis & Hippocampus  \\
& 2D CNN & Zhang \etal~2015~\cite{zhang2015deep} & N3\newline Skull stripping (LABEL)  & -  & Tissue  \\
& 3D CNN & Milletari \etal~2016~\cite{milletari2016hough}  & - & - & Deep brain structure  \\
& 3D CNN & Dou \etal~2016~\cite{dou2016automatic} & Normalisation & Connected component analysis  & Cerebral micro-bleed  \\
& 3D CNN & Urban \etal~2016~\cite{urban2014multi} & Normalisation & Connected component analysis  & Tumour  \\
& 3D CNN & Brosch \etal~2016~\cite{brosch2016deep} & Normalisation\newline Registration\newline Skull-stripping & - & MS lesions  \\
& 2D CNN & Pereira \etal~2016~\cite{pereira2016brain} &  Normalisation\newline N4ITK & Connected component analysis & Tumour  \\
& 3D CNN & Kleesiek \etal~2016~\cite{kleesiek2016deep} & Bias field correction\newline Resampling\newline Intensity scaling & Connected component analysis & Skull stripping \\
& 3D CNN & Wachinger \etal~2017~\cite{Wachinger2017} & -- & 3D CRF & Sub-cortical structure  \\
& 3D CNN & Li \etal~2017~\cite{li2017compactness} & Normalisation & -- & Deep brain structure\\
\hline
\multirow{21}{*}{\rotatebox[origin=c]{90}{Multi-path CNN}}
& 2.5D CNN & Lyksborg~\etal~2015~\cite{lyksborg2015ensemble} &  N4ITK\newline Intensity scaling &  Grow cut  & Tumour  \\
& 2.5D CNN & Birenbaum and Greenspan~\etal~2016~\cite{birenbaum2016longitudinal} &  Histogram matching &  --  & MS lesions  \\
& 2D \& 3D CNN & Brebisson~\etal~2015~\cite{de2015deep} &   -  & -  & Deep brain structure  \\
& 3D RNN & Stollenga \etal~\cite{stollenga2015parallel} & Normalisation & -- & Tissue \\
& 2D CNN &  Havaei~\etal~2016~\cite{havaei2016brain} & Outlier removal\newline Normalisation\newline N4ITK  & Connected component analysis & Tumour  \\
& 2D CNN & Zhao and Jia 2016~\cite{zhao2016multiscale} & - & -  & Tumour  \\
& 2D CNN & Moeskops \etal~2016~\cite{moeskops2016automatic} & Intensity scaling\newline Bias field correction (\cite{likar2001retrospective})\newline Skull stripping (BET) & -  & Tissue  \\
& 3D FCNN & Kamnitsas \etal~2016~\cite{kamnitsas2016efficient} & Normalisation\newline Registration\newline Skull-stripping & Fully connected 3D CRF & Tumour \\
& 2D FCNN & Shakeri \etal~2016~\cite{shakeri2016sub} & -- & 2D CRF & Sub-cortical structure \\
& 3D CNN & Chen \etal~2017~\cite{chen2016voxresnet} &  Intensity normalisation & -  & Tissue \\
& 2D \& 3D CNN & Mehta \etal~2017~\cite{mehta2017brainsegnet} & N4ITK\newline Skull stripping (BET) & -- & Sub-cortical structure \\
& 3D FCNN & Dolz \etal~2017~\cite{Dolz2017} & Bias field correction\newline Normalisation\newline Skull stripping (BET) & Connected component analysis & Sub-cortical structure \\
& 3D CNN & Valverde \etal~2017~\cite{valverde2017improving} & Normalisation & -- & MS lesions\\
\hline
\end{tabular}}
\end{table*}

\subsubsection{Interconnected operating modules}
According to the number of interconnected operating modules, the strategies in the literature can be classified into single-path and multi-path architectures. The single-path architectures correspond to the cases in which there is a unique flow of information: the input data is processed by convolutional, pooling and non-linear rectifier layers; the feature maps are then mined in the FC layers, and afterwards used for predicting the label in the output layer. An example of 2D single-path architecture is shown in Fig.~\ref{fig:2DCNN}. This category was popular in the literature~\cite{maleki2012diagnosis, dou2016automatic,pereira2016brain,kleesiek2016deep,Wachinger2017,brosch2015deep, zikic2014segmentation,dvorak2015structured} perhaps due to its fast computation and simplicity compared to multi-path architectures. In contrary to single-path networks, the multi-path architectures are a type of CNN in which independent operative networks are integrated into a single model to capture a more varied set of features. The modules can be arranged either in parallel or series. In the former approach, the idea is to achieve a consented label between different sources of information while in the latter approach the key is to reprocess the output of one network using another one. 

\begin{figure*}[!tbp]
    \centering
    \begin{subfigure}[b]{\textwidth}
        \centering
        \includegraphics[width=0.8\textwidth]{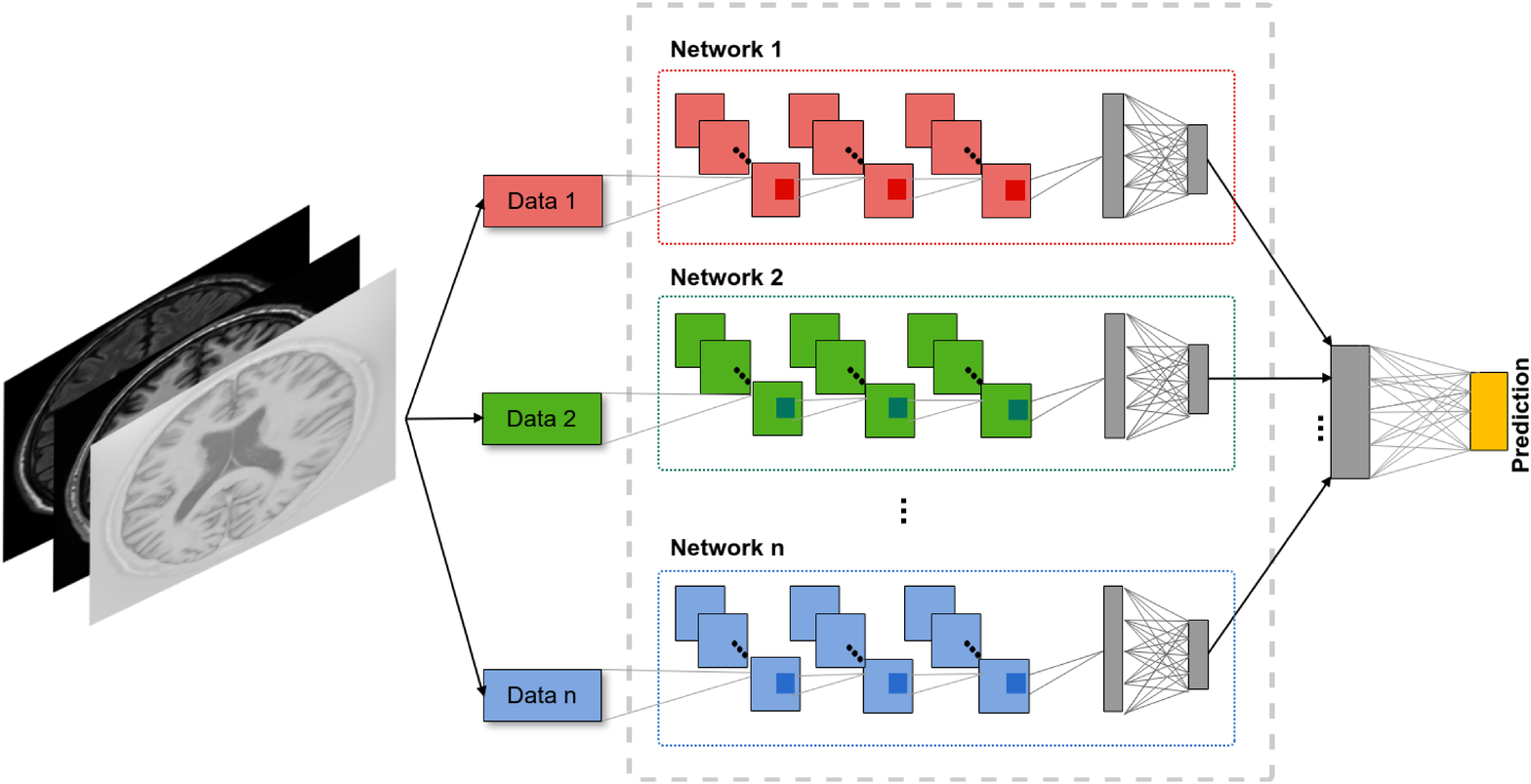}
        \caption{Multi-path architecture}
        \label{fig:mutli-path}
    \end{subfigure}\\
    \begin{subfigure}[b]{\textwidth}
        \centering
        \includegraphics[width=\textwidth]{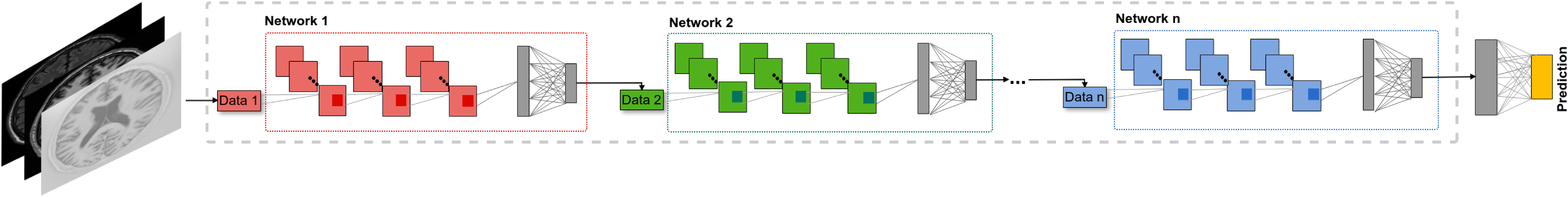}
        \vspace{-1cm}
        \caption{Cascade architecture}
        \label{fig:cascade}
    \end{subfigure}
    \caption{Architecture of multi-path CNNs. The diagram in (a) represents a multi-path CNN architecture: different operating CNNs process in parallel input data and output a consented label. The diagram in (b) corresponds to a cascade architecture of CNN networks: the output of one network is provided as input to another one to refine the segmentation.}
\end{figure*}

The parallel multi-path architectures are composed of different CNNs designed to operate in parallel to capture more comprehensive features. Each network uses different versions of the same target area. There might be two possible variants of this type of architecture. The first variant aggregates features extracted from separate networks before making predictions, which can be understood as a process of information fusion that could occur at two levels: first, at a data level, where feature maps are merged to form a larger feature map; and second, at a feature level, where the extracted final features are concatenated before making the prediction. Fig.~\ref{fig:mutli-path} graphically depicts a multi-path architecture in which different features are processed using separate networks and are concatenated before making the prediction. This concatenation could also be performed at the convolutional layers, where feature maps from different models form a larger feature map in its third dimension. The second variant of multi-path architectures consists of learning ensembles of more than two networks to improve prediction. Predictions made by each module for a given target voxel are merged probabilistically. 

There are several variants of the parallel multi-path category within the literature (i) multi-scale/multi-resolution~\cite{kamnitsas2016efficient,moeskops2016automatic, Dolz2017, urban2014multi, zhao2016multiscale} -- in which different views of the same ROI are used as input -- (ii) 2.5D~\cite{lyksborg2015ensemble} -- in which the data is extracted from the three anatomical planes -- and (iii) 2.5D+3D~\cite{de2015deep, mehta2017brainsegnet} -- in which the different modules process 2.5D and 3D patches and operate simultaneously to provide consented labels. The fusion of various information sources allows the network to obtain contrast, local detailed and implicit contextual information.

The authors in~\cite{zhao2016multiscale} and~\cite{moeskops2016automatic} used multi-scale (i.e., different views of the same target area) 2D CNNs using three separate networks and combining the features immediately before the prediction step. According to the authors, the multi-scale design allowed the network to obtain fine-grained local and general information -- presumably spatial context as well -- to produce the classification of the target pixel. Similarly, Lyksborg \etal~\cite{lyksborg2015ensemble} trained three networks separately, using 2D slices from axial, sagittal and coronal planes. Unlike in~\cite{zhao2016multiscale} and~\cite{moeskops2016automatic}, the ensemble of the three networks was performed after the prediction step, and it was based on majority vote. The ensemble, in this case, was used twice: to segment the tumour area and to perform a within-tumour segmentation. 

Another work, presented by Kamnitsas \etal~\cite{kamnitsas2016efficient}, proposed a network called DeepMedic that used a small $3 \times 3\times 3$ kernel to increase the depth of the network to 11 layers, and they used two scales to train two separate networks for brain lesion segmentation. The output of each network was fused on the fully connected layer. The 11 layer 3D CNN produced a soft segmentation map. They extended a conditional random field~\cite{krahenbuhl2011efficient} to a fully connected 3D CRF and used it as post-processing to impose a regularisation constraint and produce the final hard segmentation labels.

Brebisson and Montana~\cite{de2015deep} were able to train eight networks arranged in parallel, with each network having a seven-layer CNN architecture for whole brain anatomical segmentation. The inputs of the network captured information at different scales around the voxel of interest: 3D and orthogonal 2D intensity patches captured a local spatial context, while downscaled, large 2D orthogonal patches and the distances to the regional centroids enforced global spatial consistency. They reported over-fitting, mainly due to the enormous number of parameters of the overall model and the few training images provided in the MICCAI 2012 challenge, which was a multi-atlas labelling challenge.

The second variant of multi-path architectures is the series arrangement. In this variant, different CNN networks are cascaded in series, with information from the previous network used as input for the latter, as shown in Fig.~\ref{fig:cascade}. It also renders the overall CNN deeper and, as a result, increases the expressing power of the model. 

Havaei \etal~\cite{havaei2016brain} trained a cascaded architecture, starting by training a two-path CNN. Then, they fixed the parameters of these two-path networks and used the already trained models in the cascaded architecture. They evaluated the performance of concatenating feature maps from the last layer of the first network as feature maps for the second network at three different layers. From their three architectures submitted to the BRATS 2013 brain tumour challenge, they reported the best performance by concatenating the last layer feature maps of the first network, with the actual inputs used as input for the second network.

Valverde \etal~\cite{valverde2017improving} presented a cascaded 3D CNN approach for addressing MS lesion segmentation.  Due to the conditions of the scenario (i.e. lack of a large number of samples and data imbalance), the authors opted for performing the segmentation in two steps. In the first step, the network is trained using all the positive samples and the same number of negative samples (i.e. the negative class is under-sampled). Once the classification is performed, the probabilities for each voxel to belong to the positive class are obtained. In the second step, the network is trained with all the positive voxels and the same number of misclassified negative samples. In the paper, no details regarding the improvement of the cascade approach over the single-path architecture are quantitatively discussed. Nevertheless, with this cascaded architecture, the authors were able to outperform the rest of participants on the MICCAI 2008 challenge. 

Chen \etal~\cite{chen2016voxresnet} proposed a 3D CNN, VoxResNet, for brain tissue segmentation. The network is a 3D extension of the ResNet network~\cite{he2015deep} with a shallower depth. The authors also integrated segmentation maps from VoxResNet (used as the context information) together with the original volumes (i.e. appearance information) to train a new classifier called Auto-context VoxResNet. In this sense, possibly misclassified voxels are reprocessed and probably refined. It is important to understand that more than an ``auto-context" technique, the procedure corresponds to a cascaded CNN similar to the ones described in~\cite{havaei2016brain,valverde2017improving}. Although this classifier was intended to refine the segmentation iteratively, according to the authors, the improvement was marginal and, hence, a single iteration was considered.

In general, training multiple networks, rather than a single network, could lead to better performance. The networks can be integrated in parallel or series. The former approach permits the incorporation of different pieces of information, which help in labelling the central pixel. Different networks, which extract various useful features, could be aggregated to perform the desired segmentation. Each network arranged in parallel might need different input data. Training all of the networks arranged in parallel at the same time might demand much memory. Also, managing individual networks, training processes, selecting the best hyper-parameters and avoiding over-fitting also require a significant amount of time. The latter approximation, models arranged in series, allows refining the output of one network using another one, which needs a careful design and training process. This type of architecture performs better than single networks~\cite{havaei2016brain}.

\subsubsection{Input modalities}
The strategies in the literature can also be grouped according to the number of modalities that are processed at the same time. The categories are two: single- and multi-modality. The former case is certainly more adaptable to different scenarios as a single modality is commonly provided in datasets for tissue and sub-cortical structure segmentation (mainly T1). The latter case contemplates processing different sources of information. As stated by Zhang \etal~\cite{zhang2015deep}, it is important to consider fusing information from different imaging modalities -- if possible -- since it has been proven beneficial to discriminate when lacking contrast among tissues. The imaging sequences can be processed in parallel using multi-path configurations (e.g. one branch processes T1 and another one, FLAIR) or can be appended altogether to have multiple channels~\cite{zhang2015deep,chen2016voxresnet,lyksborg2015ensemble}. The difference between these two approaches is that specific kernels process each modality in the former case but, at the same time, the number of parameters increases.

\subsubsection{Patch dimension}
As mentioned previously, the different architectures can also be classified according to the dimension of the input patches. The categories, in this case, are three: 2D, 2.5D and 3D. 

Two-dimensional architectures consider patches from a single plane (i.e. axial, sagittal or coronal). This architecture is widely used in the literature since it is adaptable to different image domains and segmentation tasks. In semantic computer vision, these types of networks are widespread. Compared to the earlier proposal from AlexNet~\cite{krizhevsky2012imagenet}, the depth of the recently proposed networks has shown a significant increase~\cite{he2015deep,szegedy2016inception,szegedy2015going,simonyan2014very}. In MRI brain analysis, the architecture remains very similar despite being shallower, mainly because of limited available training data. It also seems unnecessary to design a very deep network if it does not improve the performance and, perhaps more importantly, it can be detrimental, due to potential over-fitting if too many network weights are required to be adjusted.

Diagnosing abnormalities employing a binary classifier is mostly the first step toward the localisation and segmentation of regions of interest. To the best of our knowledge, pioneering work using deep learning for brain image analysis was first reported in the detection of MS lesions by Maleki \etal~\cite{maleki2012diagnosis}. They used a conventional CNN architecture to extract features from 2D MR FLAIR images and a three-layer neural network as a classifier. A similar classification algorithm principle could be extended to patch-wise image segmentation algorithms. Usually incorporating multi-modal images during segmentation as a stack of 2D images improves performance. Zikic \etal~\cite{zikic2014segmentation} used 2D slices from T1, T1c, T2 and FLAIR images to classify small $19\times 19$ patches into five classes in the BRATS 2013 dataset. Their simplified CNN architecture consisted of 2D convolution, ReLU, max-pooling and softmax classifier layers.

Pixel-based labelling schemes that predict the label for a single voxel at a time have been a popular choice for training MR images using CNNs. Despite the success of this approach, scanning each voxel using a sliding window renders it computationally slow. To overcome this limitation, Dvorak and Menze \cite{dvorak2015structured} fed their network several versions of the whole 2D slice input image, shifted on the X and Y axes, and they merged the outputs accordingly. They used images from multi-modal volumes to train a CNN for the prediction of extended label patches.
   
As mentioned earlier, in the context of brain MRI, building a very deep network is a rare practice because of the difficulty of obtaining sufficient representative and accurate data for training. In spite of the performance advantages, deeper networks usually have a larger number of parameters, and they may tend to suffer from over-fitting. Training on large datasets with proper regularisation algorithms is a common practice in the community. Pereira \etal~\cite{pereira2016brain} proposed a deep architecture, eleven layers, with 2D patches extracted from T1, T1c, T2 and FLAIR for gliomas segmentation. They made use of small $ 3\times 3$ kernels which allowed designing a deeper architecture~\cite{simonyan2014very}. A CNN built using those small kernels has fewer parameters, and therefore reduces the over-fitting problem. Moreover, they also took advantage of data augmentation using rotation to increase their original training dataset. Their 2D CNN consisted of multiple 2D convolutions, followed by LeakyReLU and max-pooling layers.
 
In the reviewed literature, we observed that the conventional CNN models are popular choices of MRI analysis researchers, perhaps because of the authors' interest in utilising the already developed 2D CNN architectures, in addition to reducing computational complexity. This approach is in contrast to the actual 3D nature of the volume. 

Tri-planar or 2.5D architectures are provided with patches from the three anatomical planes (i.e. axial, sagittal and coronal), commonly using a multi-path design. As explained by Lyksborg \etal~\cite{lyksborg2015ensemble}, the 2.5D information provides a better understanding of the 3D scenario than 2D-based networks since it exploits the 3D nature of MR images and, consequently, brings up contextual information. Moreover, although 2.5D architectures are more computationally expensive than 2D variants, these networks still consider 2D convolutions and, hence, are expected to be less costly than 3D alternatives~\cite{dvorak2015structured,lai2015deep}. 

Lyksborg \etal~\cite{lyksborg2015ensemble} presented a 2.5D multi-path approach in which the patches from different planes were processed independently by similar operative modules. By using probabilistic methods, the knowledge coming from the three branches was combined after the prediction layer. Although the approach is reliable, it somehow assumes that the contribution among the three branches is the same, which may not be the case at some point (i.e. due to the anisotropic nature of the data and/or to the fact that regions of interest may be more visible in specific planes). A workaround to this situation is to perform the merging step before the prediction layer and, hence, the weight of the features to be merged are specifically fixed for the case of study since the network determines the best configuration based on their contribution to the final segmentation.

\begin{figure*}    
\centering
\includegraphics[width=\textwidth]{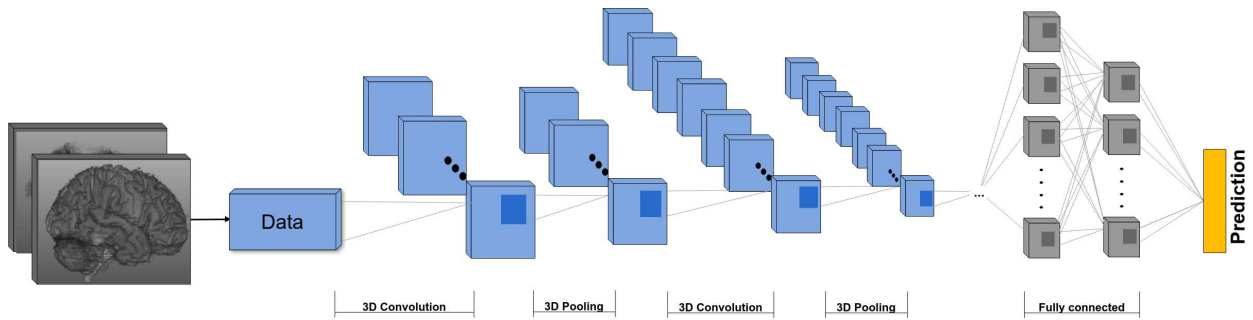}
\caption[]{Architecture of 3D CNNs. The convolution operation is performed by employing 3D filters. The pooling operation is also performed in three dimensions.\label{fig:3DCNN}}
\end{figure*}

Three-dimension architectures consist in taking 3D segments directly from the MRI volume. This architecture utilises 3D convolution kernels, which seems to be a more appropriate solution for fully exploiting the spatial contextual information in volumetric data. The main constraint of a 3D CNN approach lies in its expensive computational cost, memory requirements and computational time. Utilising the whole volume at a time is computationally very expensive; therefore, small, three-dimensional patches are extracted for training and testing. For instance, Urban \etal~\cite{urban2014multi} used four (multi-modal) $ 9^{3} $ voxels as input to segment a tumour in the BRATS 2013 dataset. Figure \ref{fig:3DCNN} shows a general 3D CNN architecture, in which the input is volumetric data, a 3D brain MR image in this case, and 3D kernels are used to perform the convolution.

Lai and Rueckert \cite{lai2015deep} investigated the use of three different convolutional network architectures for patch-based segmentation in ADNI hippocampus MRI dataset. They evaluated stacked 2D patches (one around the voxel in question, one in parallel and above it, and one in parallel and below it), tri-planar patches and 3D patches. According to their experiments, 3D convolutional architectures performed the best despite being the most computationally expensive.

To summarise, most of the algorithms, 10 of the 27, reviewed in the literature used a 2D CNN architecture. They were based on the two-dimensional convolutions used extensively in computer vision applications on natural images. This fact eases the model development process for practitioners by customising the popular CNN architectures and exploiting the available frameworks. Although patch-wise implementation of these types of architectures is very slow, converting the last output layer into a convolutional layer helps to expedite the inference time~\cite{havaei2016brain}. An implementation using a single network might not result in good performance, demonstrating that deeper networks that use smaller kernels achieve better performance, compared to using larger kernels and shallower architecture~\cite{pereira2016brain}. Two-dimensional CNNs disallow for exploiting of the actual 3D nature of the MRI data. In this sense, 2.5D and 3D CNNs were introduced mainly to overcome this limitation. Although both 2.5D and 3D architectures require more processing than 2D ones, the former type is less computationally expensive than the second one: 2D convolutional layers are used in 2.5D while 3D convolutional are used in 3D. Indeed, the main impediments to working with 3D CNNs are the number of parameters, memory requirements, expensive computational costs and slow speed. However, with the rapid development of GPUs, this situation may not represent a drawback in the coming years.

\subsubsection{Predictions at a time}
The approaches can also be grouped according to the number of predictions they perform at a time into CNN and FCNN~\cite{Dolz2017,kamnitsas2016efficient,shakeri2016sub}. The former case corresponds to the traditional approach in which a single patch is processed by the network, and a single output is returned. Since the idea at the end is to obtain segmented areas, this approach can be slow in practice. The latter case corresponds to the architectures for which the fully connected layers are replaced by $1\times 1$-kernel (or $1\times 1\times 1$-kernel) convolutional layers to obtain a dense prediction. Additionally, some approaches consider upsampling layers to increase the number of predictions at a time. Naturally, the main drawback of these approaches is that a larger number of parameters are required to be set up (hence, more training samples to train the network properly are needed) in comparison to CNN strategies. 

Dolz \etal~\cite{Dolz2017} implemented a 3D FCNN architecture which was used to segment sub-cortical structures. Apart from the FCNN implementation, the authors considered multi-resolution information by extracting feature maps from high-resolution layers and merging them with low-resolution information, coming from the main information flow, before the $1\times 1\times 1$-kernel convolutional layers. As explained by Bharath~\cite{hariharan2015hypercolumns}, this kind of connections allowed the network to learn from semantic -- coming from deeper layers -- as well as fine-grained localisation information -- coming from shallow layers.

Ronneberger \emph{et al.}~\cite{ronneberger2015u} proposed a multi-path 2D FCNN architecture called the U-Net. The network was successfully tested on the 2D electron microscopy~\cite{em2015} and ISBI 2015 cell tracking\footnote{The challenge web page is http://www.codesolorzano.com/Challenges/CTC/Welcome.html} challenges, achieving the first position in both of them. There were two key components on this FCNN. First, up-sampling layers were used to increase the resolution of the output. Thus, the volumes could be rapidly processed. In the paper, the authors were able to produce an output map of $388\times 388$ from an input image of size $572\times 572$. Second, localisation and context were retained in the network by merging features maps from high-resolution layers with the up-sampled ones. On the other hand, it is important to highlight that this type of network, as stated by its authors, requires setting up a large number of parameters. Indeed, the original implementation required data augmentation techniques (for example, non-linear deformations generated on-the-fly during training) to achieve desired segmentation results. 

Several approaches in different scenarios have been inspired by the U-Net architecture. Milletari \etal~\cite{milletari2016v} and Yu \etal~\cite{yu2017} considered a 3D extension of the U-Net along with skip and/or residual connections for addressing the MICCAI Prostate MR Image Segmentation
(PROMISE12). According to their experiments, the use of links improved the accuracy of the methods and also proved the effectiveness of information propagation. Also, the 3D variant has been implemented for confocal microscopic data~\cite{cciccek20163d} and liver and lesion segmentation in CT images~\cite{christ2016automatic}. Although none of these papers is directly applied to brain MRI, we consider that they could achieve relevance for brain MRI segmentation as well.

Brosch \etal~\cite{brosch2016deep} used 3D convolutional encoder networks for MS lesion segmentation. In addition to learning deconvolution and un-pooling, they introduced short-cut connections from the first to the last deconvolution layers inspired by the U-Net architecture~\cite{ronneberger2015u}. The network in the convolutional encoder path was initialised using weights pre-trained on a stack of convolutional restricted Boltzmann machines. With this approach, they achieved top performance in public MS lesion segmentation challenge datasets. 

\subsubsection{Contextual information}
The fact that the discussed architectures process patches separately and not the whole volume has some implications. For instance, the spatial distribution of brain structures is not directly encoded. Although some implicit contextual information is encoded in 2D, 2.5D and 3D patches their information is limited to the size of the patches. The bigger the size of the patch, the more information the network can take into account to produce the prediction; but the more the parameters to be trained. When considering the architecture of Havaei \etal~\cite{havaei2016brain} and the work of Kamnitsas \etal~\cite{kamnitsas2016efficient}, it seems to show that the context is important, but to preserve spatial correspondence of the multi-scale features is crucial. This could explain why a simple approach as \cite{pereira2016brain} could outperform~\cite{havaei2016brain}.

Recently, Wachinger \etal~\cite{Wachinger2017} introduced explicit within-brain location information through Cartesian and spectral coordinates~\cite{Spectral2013, Cartesian2014} to help the classifier to discriminate one class from another. The Cartesian coordinates were obtained by taking the XYZ-coordinates of each of the voxels while the spectral coordinates came from the eigenvectors of the Laplacian matrix of the graph representing the brain mask. As explained by the authors in~\cite{Wachinger2016}, the explicit spatial information is useful to (i) discriminate between patches from different brain hemispheres since patch-based strategies loss spatial context and (ii) help in the segmentation of tissues exhibiting low contrast. In this latter approach, it is essential to have in mind that (i) considering XYZ-coordinates on the image plane makes sense only if the volumes are registered and (ii) the distribution of the eigenvectors depends highly on the shape of the brain mask, i.e. these features may not be reliable when the shape of the region of interest differs considerably from one volume to another.

\subsubsection{Summary of pros and cons of the different strategies}
Table \ref{tab:proscons_summary} summarises the main advantages and disadvantages of the approaches discussed previously in this section. 

\begin{table*}[ht!]
\centering
 \caption{Pros and cons of the segmentation strategies presented. \label{tab:proscons_summary} }
     \resizebox{\textwidth}{!}{ 
    \begin{tabular}{l p{8cm} p{8cm}}
      \hline
      \textbf{Strategy} & \textbf{Advantage} & \textbf{Disadvantage}\\
      \hline \hline
      \multicolumn{3}{c}{\textit{\textbf{Interconnected operating modules}}} \\ \hline
      Single path & In principle, fast computation & Single flow of information\\ \hline
      Multi-path - parallel & Extract more varied features\newline Verdict is consented among interconnected modules\newline Incorporated information may provide the network with contextual information (e.g. multi-resolution architectures) & More parameters than single-path architectures\newline More computationally demanding (data preparation and processing)\\ \hline
      Multi-path - series & Extract more varied features\newline Allows refining information at different stages & Requires careful design of the network\newline May require training different networks\newline The improvement could be marginal\\ \hline \hline
      \multicolumn{3}{c}{\textit{\textbf{Input modalities}}} \\ \hline
      Single modality & Adaptable to different scenarios\newline Easily extensible to multiple modalities (channels or multi-path) & Single source of information \\ \hline
      Multi-modality & Gain valuable contrast information & Requires more parameters than single modality \\ \hline \hline
      \multicolumn{3}{c}{\textit{\textbf{Patch dimension}}} \\ \hline
      2D & Easily scalable to complex network architectures\newline Flexibility and adaptability\newline Fast computation & Heavily dependent on the network design tricks to obtain good results\newline Disallow 3D nature of MRI\\ \hline
      2.5D & Faster computation than 3D\newline Exploits the 3D nature of MRI\newline Acquires implicit contextual information & Computationally more expensive than 2D\\ \hline
      3D CNN & Exploits 3D nature of the MRI volume directly\newline Usually leads to better performance than 2D\newline Acquires implicit contextual information & Computationally expensive\newline Scaling to larger patch size may be computationally demanding\newline Due to the number of parameters, it may require large training data\\ \hline \hline
      \multicolumn{3}{c}{\textit{\textbf{Predictions at a time}}} \\ \hline
      CNN & In principle, less number of parameters than FCNN & Single classification for a single patch\\ \hline
      FCNN & Faster segmentation (some architectures can classify a single volume in one shot) & Larger number of parameters to be set up\newline Generally, requires more training samples (data augmentation is commonly adopted in this sense)\\ \hline
      \multicolumn{3}{c}{\textit{\textbf{Contextual information}}} \\ \hline
      Implicit & No additional calculations are required & Context depends on the size of the patch \\ \hline
      Explicit & Discriminative features are provided to the network\newline Sense of within-brain positioning & Subject to registration accuracy or shape of the region of interest \\ \hline
    \end{tabular}}
\end{table*}

\subsection{Post-processing}
Post-processing is the step to refine or improve the results of the segmentation. The purpose of post-processing could be either to create hard segmentation labels~\cite{kamnitsas2016efficient} or to remove false positives~\cite{dou2016automatic,zhang2015deep,pereira2016brain,havaei2016brain,urban2014multi,lai2015deep} from segmentation results, using different algorithms such as connected component analysis and Conditional Random Fields (CRF). The former strategy consists in keeping only the $k$ largest areas from the segmentation to remove spurious outputs. Nevertheless, in sub-cortical structure segmentation these spurious areas may have a larger volume than the small structures of interest and, hence, it becomes impractical. The latter strategy, CRF model, considers not only the given voxel but also its neighbours to produce a refined label, especially around the edges. Although CRF-based methods require learning parameters, they could bring regularisation properties to the segmentation since they are based on the minimisation of an energy equation. This stage helps to improve the overall performance of the model at the expense of adding computational complexity.

\begin{figure}
    \centering
    \begin{subfigure}[b]{0.2\textwidth}
        \includegraphics[width=\textwidth]{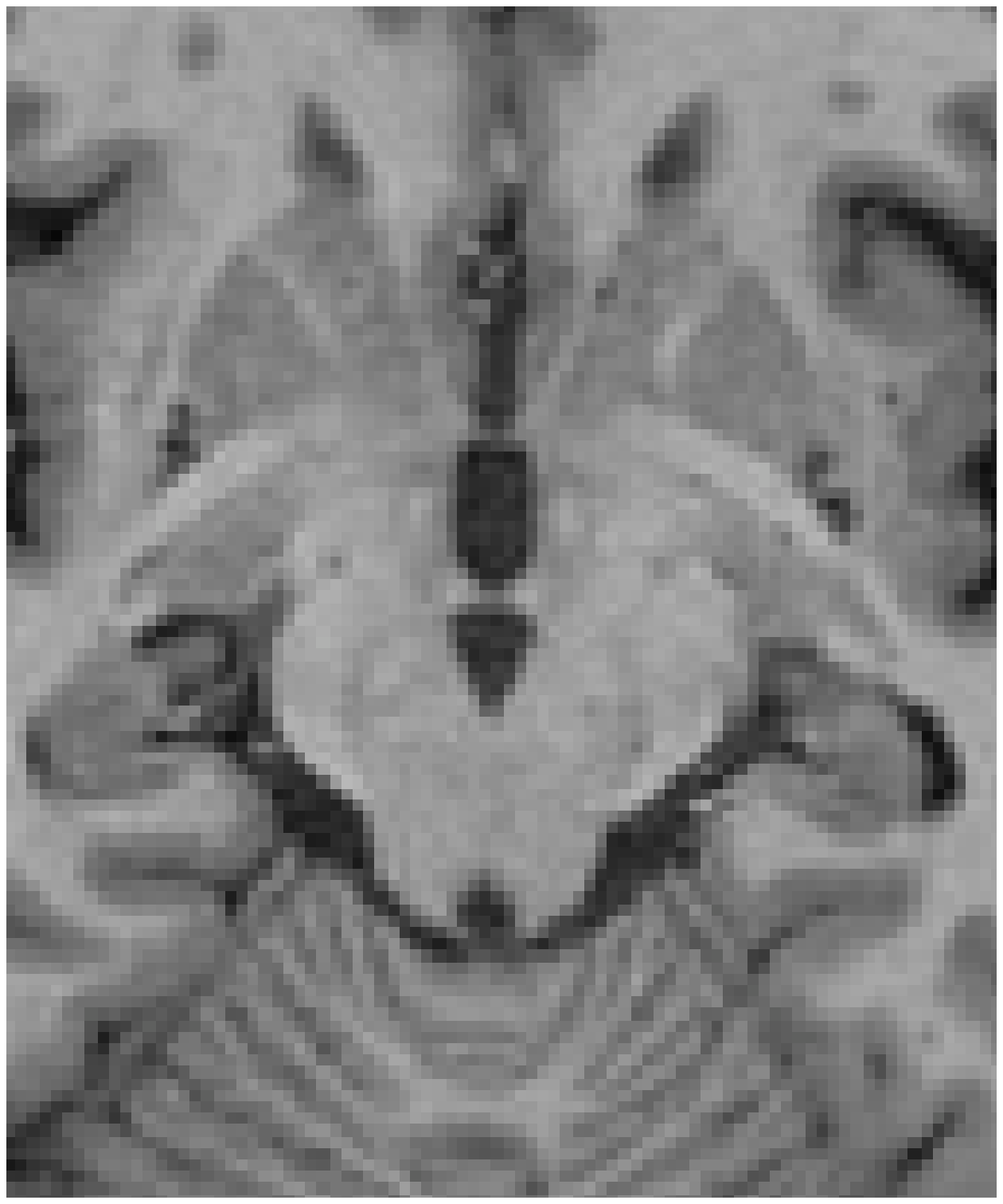}
        \caption{}
    \end{subfigure}
    ~
    \begin{subfigure}[b]{0.2\textwidth}
        \includegraphics[width=\textwidth]{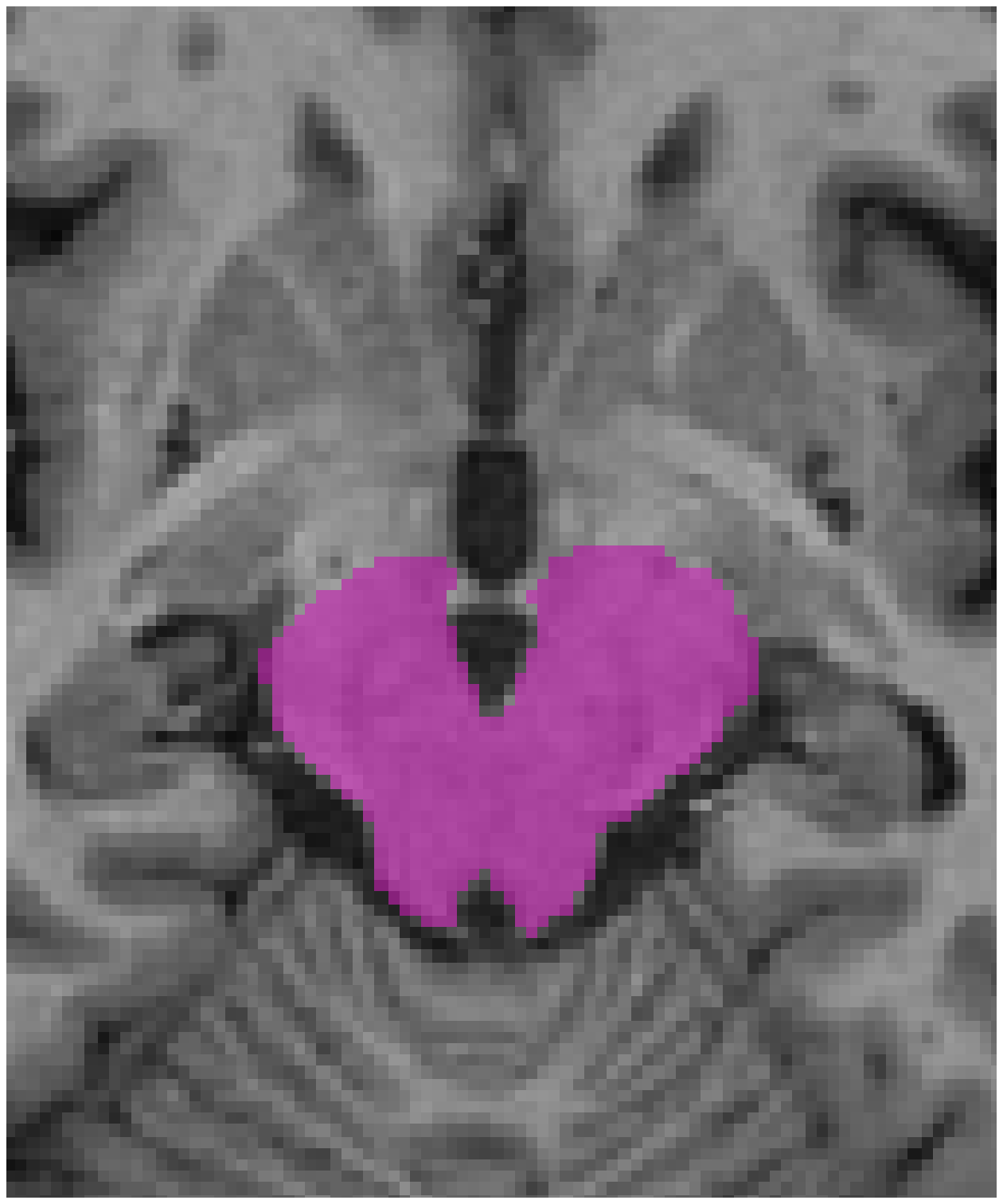}
        \caption{}
    \end{subfigure}
    ~
    \begin{subfigure}[b]{0.2\textwidth}
        \includegraphics[width=\textwidth]{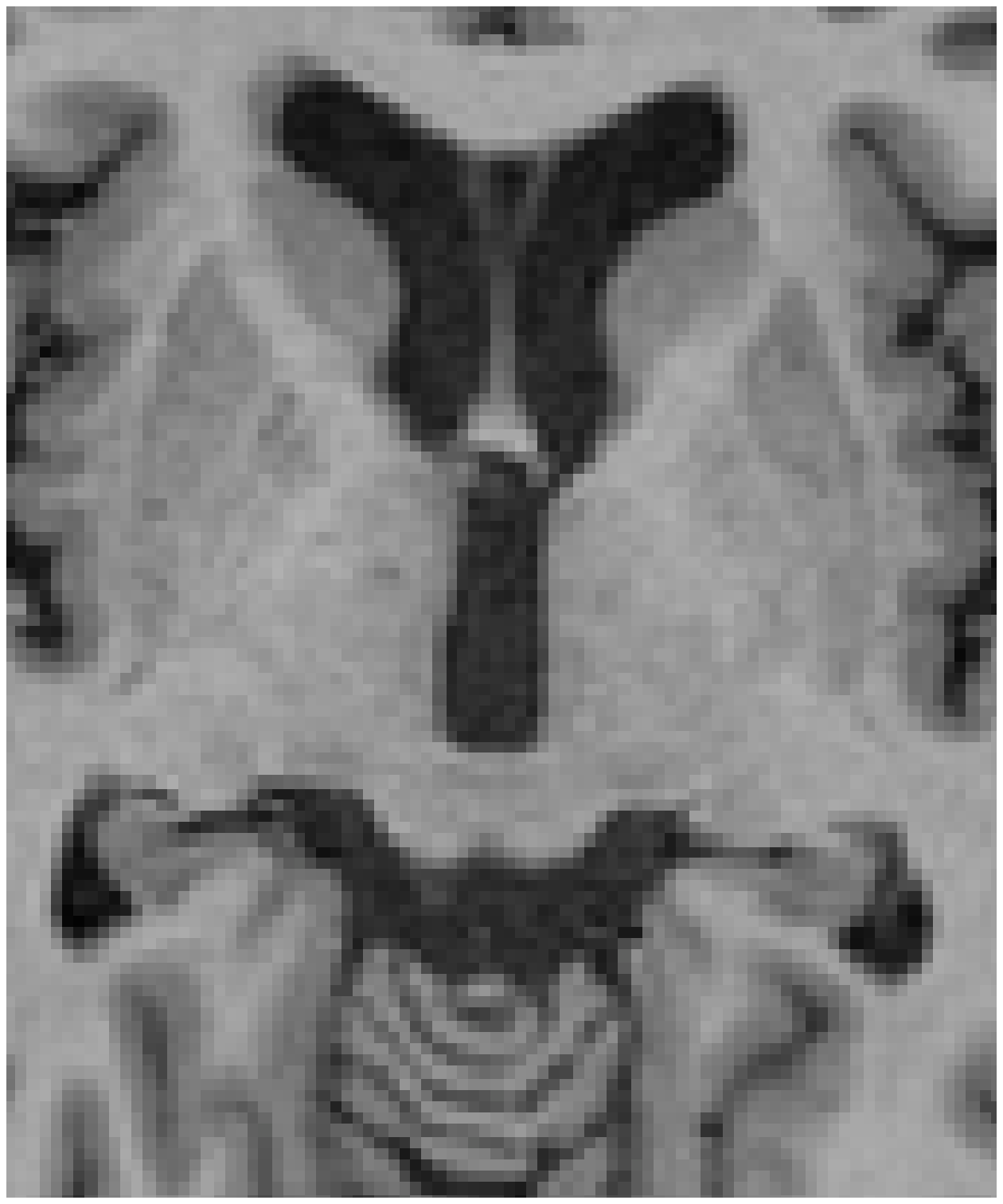}
        \caption{}
    \end{subfigure}
    ~
    \begin{subfigure}[b]{0.2\textwidth}
        \includegraphics[width=\textwidth]{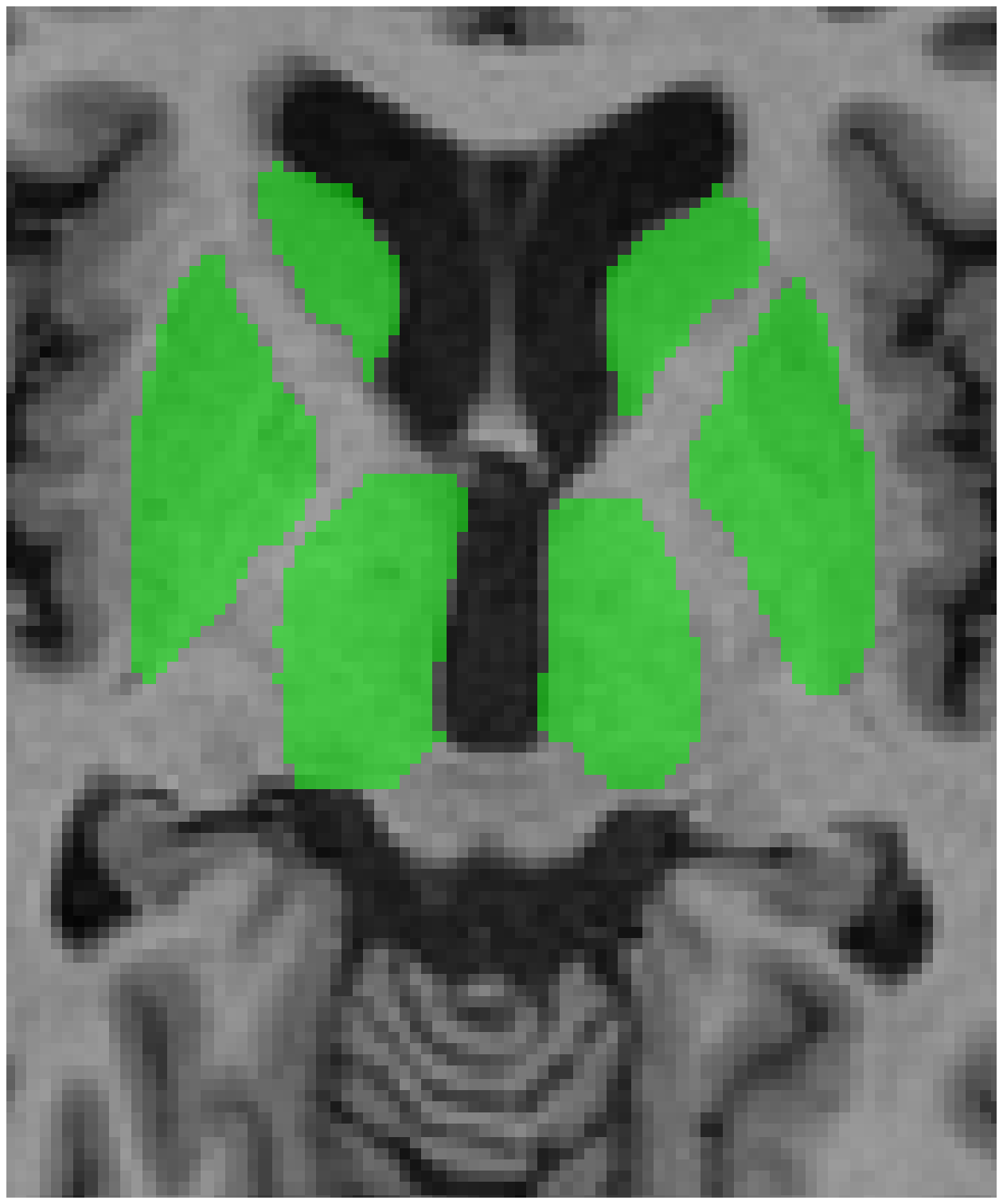}
        \caption{}
    \end{subfigure}
    \caption[Low contrast regions]{Two input slices in which low contrast between the region of interest and the surrounding areas is exhibited. In the image (a) the focus is on the mesencephalon which is highlighted by the purple area in (b). In (c) the focus is on caudate, putamen, pallidum and thalamus structures presented in green in (d).\label{fig:low-contrast}}
\end{figure}

Contrast-driven post-processing techniques may lead to unexpected/unsatisfactory smoothing results. For instance, sub-cortical structures and some parts of the brainstem could be shrunk in this step due to low contrast with surrounding areas as shown in Fig.~\ref{fig:low-contrast}. However, this does not signify that the method is not capable of refining the segmentation since both the intensity features and the prior probabilities take part in the process. 

In addition to CRF and connected component analysis, Lyksborg \etal~\cite{lyksborg2015ensemble} utilised an algorithm called Grow cut~\cite{vezhnevets2005growcut}. It was initially proposed as a continuous state cellular automata method for automated segmentation based on user labelled seed voxels. The authors used it to refine the initial tumour segmentation results. The post-processing algorithm was applied iteratively until a stable segmentation was obtained.

\section{Evaluation \label{section5}}
Brain MRI volume segmentation approaches are usually evaluated using different quantitative measurements. The described CNN models were tested in public and clinical research trial datasets. Regarding the works reviewed, both the most common datasets and the standard measurements used for the evaluation are described in this section. Moreover, the results obtained by the analysed methods on the standard datasets are presented.  

\subsection{Public evaluation frameworks and datasets}
Several tissue segmentation algorithms have been proposed during the last years in the literature. However, determining which of them achieves the best performance can be complicated in principle since not all the algorithms are publicly available. A way to address this situation is by creating standard evaluation frameworks, such as the grand challenges taking place on relevant international conferences. Thus, before delving into a detailed discussion about the methods, we describe some of the most widely used public datasets for brain lesions, structures and anatomical segmentation.

\begin{table*}[ht!]
    \caption{Summary of commonly used public datasets in MS, tumour, tissue and structure segmentation. The table is structured as follows. The different datasets are listed in the first column and their corresponding information: number of scans, offered modalities, acquisition scanner, segmentation tasks in which it is used, and image information such as reconstruction matrix and pixel spacing are detailed from column two to seven. \label{tab:dataset_summary}}
    \resizebox{\textwidth}{!}
    { 
        \begin{tabular}{lccM{3cm}M{3cm}p{2.4cm}M{2.5cm}M{3.1cm}}
            \hline
            \multirow{2}{*}{\textbf{Name}} &  \multicolumn{2}{c}{\textbf{Number of scans}} & \multirow{2}{*}{\textbf{Modality}} & \multirow{2}{*}{\textbf{Scanner}} & \multirow{2}{2.4cm}{\textbf{Segmentation tasks}} & \multirow{2}{2.5cm}{\textbf{Reconstruction matrix}} & \multirow{2}{3.1cm}{\textbf{Pixel spacing ($\mathbf{mm}$)}}\\ \cline{2-3}
            & \textbf{Train} & \textbf{Test} & & & & & \\ \hline\hline
            MICCAI 2008~\cite{styner20083d} &  20 & 24 &  T1, T2, and FLAIR  & 3T Siemens Allegra  &  MS lesion   & $ 512\times 512\times 512 $ & $ 0.5 \times 0.5 \times 0.5  $ \\\hline
            MICCAI 2012 & 15 & 20 & T1 & -- & Whole brain, tissue and sub-cortical structure & $ 512\times 512\times 512 $ & $ 0.5 \times 0.5 \times 0.5  $ \\\hline
            NeoBrainS12~\cite{ivsgum2015evaluation} & 20 & 5  & T1 and T2 & 3T Philips Achieva & Tissue & $384\times 384\times 50$\newline$512\times 512\times 110$\newline$512\times 512\times 50$ & $0.34\times 0.34\times 2.0$\newline$0.35\times 0.35\times 1.2$\newline$0.35\times 0.35\times 2.0$ \\\hline
            IBSR18 & 18 & -- & T1 & -- & Tissue and sub-cortical structure  & $256\times 128\times 256$ & $0.84\times 0.84\times 1.50$\newline$0.94\times 0.94\times 1.50$\newline$1.00\times 1.00\times 1.50$ \\ \hline
            MRBrainS13~\cite{mendrik2015mrbrains}  & 5 & 15 & T1,T1\_1mm, T1\_IR and FLAIR & 3T Philips Achieva & Tissue  & $  256 \times 256 \times 192  $\newline $  240 \times 240 \times 48  $ &  $0.958 \times 0.958 \times 3.0$ \newline T1\_1mm has  $ 1 \times 1 \times 1 $ \\\hline            
            BRATS 2012~\cite{menze2015multimodal} & 80 & 30 & T1, T1c, T2, and FLAIR & -- & Tumour & $160 \times 216 \times 176$\newline $176 \times 176 \times 216$ & $ 1\times 1 \times 1$  \\ \hline
            BRATS 2013~\cite{menze2015multimodal} & 30 & 25 & T1, T1c, T2, and FLAIR & -- & Tumour & $160 \times 216 \times 176$\newline $176 \times 176 \times 216$ & $ 1\times 1 \times 1$  \\ \hline
            BRATS 2014~\cite{menze2015multimodal} & 309 & -- & T1, T1c, T2, and FLAIR & -- & Tumour & $160 \times 216 \times 176$\newline $176 \times 176 \times 216$ & $ 1\times 1 \times 1$  \\ \hline
            BRATS 2015~\cite{menze2015multimodal} & 274 & 110 & T1, T1c, T2, and FLAIR & -- & Tumour  & $240 \times 240\times 150$ & $ 1 \times 1 \times 1 $ \\ \hline
            ISBI 2015~\cite{carass2017longitudinal} & 21 & 61 &  T1, T2, PD, and FLAIR & 3T MRI &  Longitudinal MS lesion  &  $ 181 \times 217 \times 181  $ & $1 \times 1 \times 1$ \\ \hline
            \multirow{5}{*}{ISLES 2015~\cite{maier2017isles}} & 28 & 36 &  FLAIR, T2 TSE, T1 TFE/TSE, DWI & 3T Philips & \multirow{5}{3cm}{Ischemic stroke lesion} & $230\times 230\times 154$ & $1\times 1\times 1$\\
            & 30 & 20 & T1c, T2, DWI, CBF, CBV, TTP, Tmax & 1.5T Siemens Magnetom Avanto\newline 3T Siemens Magnetom Trio & & -- & $2\times 2\times 2$  \\ \hline
            MSSEG 2016~\cite{commowick2016msseg} & 15 & 38 & T1, T1 GADO, FLAIR, DP/T2 & 3T GE Discovery\newline 3T Philips Ingenia\newline 1.5T Siemens Aera\newline 3T Siemens Verio & MS lesion & $176\times 256\times 256$\newline $256\times 256\times 176$\newline $200\times 336\times 336$\newline $210\times 336\times 336$ & $1\times 1\times 1$\newline $1.08\times 1.08\times 0.9$\newline $0.85\times 0.74\times 0.74$ \\ \hline
            LBPA40 & 40 & -- & T1 & 1.5T GE & Brain structure & $181\times 217\times 181$ & $1\times 1\times 1$\\ \hline
            Hammers95n30 & 30 & -- & T1 & 1T Philips HPQ & Brain structure & Varied & $0.94\times 0.94\times 0.94$\\ \hline
            Hammers83n30 & 30 & -- & T1 & 1T Philips HPQ & Brain structure & Varied & $0.94\times 0.94\times 0.94$\\ \hline
            Hammers67n20 & 20 & -- & T1 & 1T Philips HPQ & Brain structure & Varied & $0.94\times 0.94\times 0.94$\\ \hline
        \end{tabular}}
\end{table*}
  
Some of the most commonly used public datasets in brain image MRI analysis are summarised in Table~\ref{tab:dataset_summary}. The reviewed works primarily used MICCAI datasets for MS lesions, tumours, tissue and structure segmentation. 
For MS lesion segmentation, the challenge was created in conjunction with MICCAI 2008\footnote{\url{http://www.ia.unc.edu/MSseg/index.html}}. The challenge was part of the 3D segmentation in the Clinic Grand Challenge II and consisted of 54 brain MRI images, 20 of which are available for training with their corresponding ground truth. For brain tumour segmentation, the series of multi-modal brain tumour segmentation (BRATS) challenges are commonly used. These challenges started in 2012 with 80 cases of real and synthetic data and, every year, the size of the training and testing data has been enlarged (up to 300+ cases in total). For sub-cortical structure segmentation, the MICCAI 2012 \footnote{\url{https://masi.vuse.vanderbilt.edu/workshop2012/index.php/Challenge_Details}} challenge in multi-atlas labelling and the Internet Brain Segmentation Repository (IBSR)\footnote{Available at http://www.nitrc.org/projects/ibsr} are widely used. It is important to note that MICCAI 2012 challenge is initially intended to evaluate algorithms on whole-brain structure segmentation, but naturally, the labels can be ignored and merged to obtain the sub-cortical structures only. Additionally, the IBSR dataset can be used for both sub-cortical structures and brain tissue segmentation. Other datasets widely adopted to evaluate algorithms for tissue segmentation are two the MICCAI Grand Challenges MRBrainS13\footnote{\url{http://mrbrains13.isi.uu.nl/index.php}} and NeoBrainS12\footnote{\url{http://neobrains12.isi.uu.nl }}. The difference between the two datasets is that in the former case, the algorithms are evaluated using adults exhibiting WM lesions and, in the latter case, the data comes from neonatal subjects.

The main evaluation measures for the challenges mentioned previously are DSC, specificity, sensitivity, positive predictive value (precision), average surface distance (ASD), average volumetric difference (AVD) and modified Hausdorff distance (MHD). 

\subsection{Evaluation measurements}
Regarding the application, we found that the CNN models were applied for MS lesion segmentation, brain tumour segmentation and structure segmentation. Although there are public datasets available for these applications, researchers still prefer to work on their databases. One main reason might be the limitation of the number of training samples available in the public datasets. As stated before, working with a small dataset affects the performance of deep CNN models enormously and imposes restrictions on practitioners from fully exploiting the capacity of those algorithms. 

The evaluation measurements compare the output of segmentation algorithms with ground truth in either a pixel-wise or a volume-wise basis. For a more robust assessment, more than one expert might be involved in generating the ground truth volumes to avoid inter-observer variability. In the reviewed literature, the results were evaluated using different evaluation metrics. In cases of MS and tumour segmentation, the most common evaluation measurements are DSC, precision, recall, the true positive rate (TPR) and the positive predictive value (PPV). Also, absolute volume difference, lesion-wise true positive rate (LTPR) and the lesion-wise false positive rate (LFPR) are also employed. Similarly, for brain tumour and structure segmentation applications, the DSC score, precision, and recall are among the widely used measurements. One main reason for this choice is that the challenge organisers primarily rely on these results. Table~\ref{tab:evaluation_summary} summarises the types of databases, numbers of samples, modalities considered, evaluation measurements applied and corresponding results reported of the surveyed works.

\begin{center}
{\tiny
\bgroup
\def\arraystretch{1.0}
\begin{longtable}{l l ccc p{2cm} M{0.8cm} p{1.3cm} p{3.5cm}}
\caption{Summary of results in the reviewed papers. The acronyms for the sequences stand for: susceptibility-weighted imaging (SWI), quantitative susceptibility mapping (QSM), proton density (PD), fractional anisotropy (FA), magnetisation prepared rapid gradient echo (MPRAGE) and gradient echo (GE). The acronyms in the measurements stand for: modified Hausdorff distance (MHD), positive predictive (PP) and average symmetric surface distance (ASSD). The acronyms in results are: grey matter (GM), white matter (WM), cerebrospinal fuild (CSF), dataset from University of North Carolina (UNC) and dataset from Children's Hospital Boston (CHB).\label{tab:evaluation_summary}}
\endfirsthead
\hline 
\multirow{2}{*} {Article}  & \multirow{2}{*}{Dataset}  & \multicolumn{3}{c}{Number of samples} & \multirow{2}{*}{Modality} & Classes & \multirow{2}{1cm}{Evaluation measures} & \multirow{2}{*}{Result}  \\ \cline{3-5}
 & & Train & Val & Test & & & \\
\hline \hline
\multicolumn{5}{l}{\textbf{Lesion}} \\ \hline
\multirow{3}{*}{Maleki \etal~\cite{maleki2012diagnosis}} & Clinical trial & 152 & - & - & FLAIR & 2 & Sensitivity\newline Specificity\newline Accuracy & 96.1\%\newline 97.5\%\newline92.9\%\\ \cline{1-9}
\multirow{11}{*}{Brosch \etal~\cite{brosch2016deep}} & MICCAI 2008 & 20 & - & 23 & T1, T2, PD and FLAIR & 2 & VD\newline TPR\newline FPR & UNC 63.5\%, CHB 52.0\%\newline UNC 47.1\%, CHB 56.0\%\newline UNC 52.7\%, CHB 49.8\%\\ \cline{2-9}
& ISBI 2015 & 20 & 1 & 61 & T1, T2, PD and FLAIR & 2 & DSC\newline LTPR\newline LFPR & 68.3\%\newline 78.3\%\newline 64.5\%\\ \cline{2-9}
& Clinical trial & 250 & 50 & 77 & T1, T2 and FLAIR & 2 & DSC\newline LTPR\newline LFPR\newline VD & 63.8\%\newline 62.5\%\newline 36.2\%\newline 32.9\%\\ \cline{1-9}
\multirow{4}{*}{Chen \etal~\cite{chen2015automatic}} & Clinic data & 55 & - & 55 & - & 2 & Sensitivity\newline Precision\newline F1-score\newline Average FP & 89.0\%\newline 56.0\%\newline 0.69\newline 6.4 \\ \cline{1-9}
\multirow{3}{*}{Dou \etal~\cite{dou2016automatic}} & Clinical data & 924 & 108 & 117 & SWI & 2 & Sensitivity\newline Precision\newline Mean FP & 93.16\%\newline 44.13\%\newline 2.74 \\ \cline{1-9}
Birenbaum and Greenspan~\cite{birenbaum2016longitudinal} & ISBI 2015 & 21 & - & 61 &  T1, T2, PD and FLAIR & 2 & DSC & 62.7\% \\ \cline{1-9}
\multirow{13}{*}{Valverde \etal~\cite{valverde2017improving}} & MICCAI 2008 & 20 & - & 23 & T1, T2, PD and FLAIR & 2 & VD\newline TPR\newline FPR & UNC 62.5\%, CHB 48.8\%\newline UNC 55.5\%, CHB 68.7\%\newline UNC 46.8\%, CHB 46.0\% \\ \cline{2-9}
& Clinical trial 1 & 35 & -- & -- & T1, T2 and FLAIR & 2 & DSC\newline VD\newline TPR\newline FPR\newline PPV & 53.5\%\newline 30.8\%\newline 77.0\%\newline 30.5\%\newline 70.3\%\\ \cline{2-9}
& Clinical trial 2 & 25 & -- & -- & T1, T2 and FLAIR & 2 & DSC\newline VD\newline TPR\newline FPR\newline PPV & 56.0\%\newline 27.5\%\newline 68.2\%\newline 33.6\%\newline 66.1\% \\ \hline\hline

\multicolumn{5}{l}{\textbf{Tumour}} \\ \hline
Zikic \etal~\cite{zikic2014segmentation} & BRATS 2013 & 20 & - & - & - & 5 & DSC & Complete 83.7\%, Core 73.6\%, Enhanced 69.0\%\\ \cline{1-9}
Urban \etal~\cite{urban2014multi} & BRATS 2013 & 30 & - & 15 & T1, T1c, T2 and FLAIR & 5 & DSC & Complete 87.0\%, Core 77.0\%, Active 73.0\% \\ \cline{1-9}
Dvorak \& Menze~\cite{dvorak2015structured} & BRATS 2014  & 166 & 25 & 66 & T1,T1c,T2 and FLAIR & & DSC & Complete 83\% , Core 75\%, Active 77\%    \\  \cline{1-9}
\multirow{3}{*}{Lyksborg \etal~\cite{lyksborg2015ensemble}} & BRATS 2014  & 91 & - & 187 & T1, T2, PD, and FLAIR & 2 & DSC\newline Positive predictive\newline Sensitivity & Complete 79.9\%, Core 63.1\%, Enhancing 62.5\%\newline Complete 0.78, Core 0.63, Enhancing 0.58\newline Complete 0.86, Core 0.74 Enhancing 0.78 \\  \cline{1-9}
\multirow{4}{*}{Pereira \etal~\cite{pereira2016brain}} & BRATS 2013 & 30 & - & 25 & T1, T1c, T2 and FLAIR & 5 & DSC\newline PPV\newline Sensitivity & Complete 88\%, Core 83\%, Enhanced 77\%\newline Complete 0.85, Core 0.82, Enhanced 0.60\newline Complete 0.86, Core 0.76, Enhanced 0.68\\ \cline{2-9}
& BRATS 2015 & 30 & - & 25 & T1, T1c, T2 and FLAIR & & DSC & Complete 78.0\%, Core 65.0\%, Enhanced 75.0\%\\ \cline{1-9}
\multirow{13}{*}{Kamnitsas \etal~\cite{kamnitsas2016efficient}} & Clinical data & 46 & - & 15 & MPRAGE, FLAIR, T2, PD and GE & 2 & DSC\newline Precision\newline Sensitivity\newline ASSD\newline HD & 64.5\%\newline 69.8\%\newline63.9\%\newline 3.72\newline 52.38 \\ \cline{2-9}
& BRATS 2015 & 274 & - & 110 & T1, T1c, T2 and FLAIR & 5 & DSC\newline Precision\newline Sensitivity & Complete 84.9\%, Core 66.7\%, Enhanced 63.4\%\newline Complete 85.3\%, Core 86.1\%, Enhanced 63.4\%\newline Complete 87.7\%, Core 60.0\%, Enhanced 67.4\%\\ \cline{2-9}
& ISLES 2015 & 28 & - & 36 & T1, T1c, FLAIR and DWI & 2 & DSC\newline Precision\newline Sensitivity\newline ASSD\newline HD & 59.0\%\newline 68.0\%\newline 60.0\%\newline 7.87\newline 39.61\\ \cline{1-9}
\multirow{3}{*}{Havaei \etal~\cite{havaei2016brain}} & BRATS 2013  &  \vtop{\hbox{\strut 30}} & - & 25 & T1, T1c,T2 and FLAIR & 5 & DSC\newline Sensitivity\newline Specificity & Complete 84.0\%, Core 71.0\%, Enhancing 57.0\%\newline Complete 0.88, Core 0.79, Enhancing 0.54\newline Complete 0.84, Core 0.72, Enhancing 0.68 \\  \cline{1-9}
Zhao and Jia \cite{zhao2016multiscale} & BRATS 2013 & 80 &  - & 30 &  T1, T1c,T2 and FLAIR & 5 & DSC & 81.0\% $\pm$ 9.5\%\\  \cline{1-9}
 \hline \hline
 
\multicolumn{5}{l}{\textbf{Tissue}} \\ \hline
\multirow{2}{*}{Zhang \etal~\cite{zhang2015deep}} & Clinic data & 7 & 1 & - & T1, T2 and FA & 3 & DSC\newline MHD & 85.03\%$\pm$ 2.27\%\newline 0.32$\pm$ 0.12 \\  \cline{1-9}

\multirow{3}{*}{Stollenga~\cite{stollenga2015parallel}} & MRBrainS13 & 5 & -- & 15 &  T1, FLAIR and T1\_IR & 5 & DSC\newline MHD\newline AVD & WM 88.33\%, GM 84.82\%, CSF 83.72\%\newline WM 2.07, GM 1.69, CSF 2.14\newline WM 7.05, GM 6.77, CSF 7.10\\ \cline{1-9}

\multirow{6}{*}{Moeskops \etal~\cite{moeskops2016automatic}} & NeoBrainS12 & - & - & - & T2 & 8 & DSC & Cor. 30 wks 82.7\%, Ax. 40 wks 80.5\%, Cor. 40 wks 81.9\%\\\cline{2-9}

& MRBrainS13 & 5 & - & 15 & T1 & 7 & Mean DSC\newline Mean ASD & 85.86\%\newline0.93 \\ \cline{2-9}
& MICCAI 2012 & 15 & - & 20 & T1 & 6 & Mean DSC\newline Mean ASD & 90.66\%\newline 0.50 \\ \cline{1-9}
\multirow{3}{*}{Chen \etal~\cite{chen2016voxresnet}} & MRBrainS13 & 5 & - & 15 & T1, T1\_IR and FLAIR & 4 & DSC\newline MHD\newline AVD & WM 89.39\%, GM 86.12\%, CSF 83.96\%\newline WM 1.94, GM 1.47, CSF 2.28\newline WM 5.84, GM 6.42, CSF 7.44\\ \hline \hline

\multicolumn{5}{l}{\textbf{Structure}} \\ \hline
Lai~\cite{lai2015deep} & ADNI & 120 & 40 & 40 & - & 3 & Error rate & 7.21\% \\ \cline{1-9}
Brebisson \etal~\cite{de2015deep} & MICCAI 2012 & 15 & - & 20 & T1 & 134 & Mean DSC & 72.5\% \\ \cline{1-9}
Milletari \etal~\cite{milletari2016hough} & Clinical data & 45 & - & 10 &  QSM & 26 & Mean DSC & 77.0\% \\ \cline{1-9}
Shakeri \etal~\cite{shakeri2016sub} & IBSR18 & 18 & -- & -- & T1 & 5 & Mean DSC & 82.40\% \\ \cline{1-9}
\multirow{5}{*}{Mehta \etal~\cite{mehta2017brainsegnet}} & MICCAI 2012 & 15 & -- & 20 & T1 & 134 & Mean DSC & 74.30\% \\ \cline{2-9}
& IBSR18 & 9 & -- & 9 & T1 & 32 & Mean DSC & 84.4\% \\ \cline{2-9}
& LONI-LPBA40 & 20 & -- & 20 & T1 & 54 & Mean DSC & 82.4\% \\ \cline{2-9}
& Hammers67n20 & 10 & -- & 10 & T1 & 67 & Mean DSC & 84.0\% \\ \cline{2-9}
& Hammers83n30 & 15 & -- & 15 & T1 & 83 & Mean DSC & 80.8\% \\ \cline{1-9}
Wachinger \etal~\cite{Wachinger2017} & MICCAI 2012 & 15 & -- & 15 & T1 & 25 & Mean DSC & 90.6\%\\ \cline{1-9}
\multirow{2}{*}{Dolz \etal~\cite{Dolz2017}} & IBSR18 & 12 & 3 & 3 & T1 & 4 & Mean DSC & 90.0\% \\ \cline{2-9}
& ABIDE & 150 & 15 & 947 & T1 & 4 & Mean DSC & 90.0\% \\ \cline{1-9}
Li \etal~\cite{li2017compactness} & ADNI & 443 & 50 & 50 & T1 & 155 & DSC & $84.34\pm 1.89$ \\ \hline

\multicolumn{5}{l}{\textbf{Skull stripping}} \\ \hline
\multirow{6}{*}{Kleesiek \etal~\cite{kleesiek2016deep}} & IBSR & 18 & -- & -- & T1 & 2 & DSC & $95.77\pm 0.01$ \\ \cline{2-7}
 & LPBA40 & 40 & -- & -- & T1 & 2 & Sensitivity & $94.25\pm 0.03$  \\ \cline{2-7}
 & OASIS & 77 & -- & -- & T1 & 2 & Specificity & $99.36\pm 0.003$ \\ \cline{2-9}
 & Clinical data & 53 & -- & -- & T1, T1c, T2, FLAIR & 2 & DSC\newline Sensitivity\newline Specificity & $95.19\pm0.01$\newline $96.25\pm 0.02$\newline $99.24\pm0.003$\\ \cline{1-9}
\end{longtable}
\egroup}
\end{center}

\section{Discussion and future directions \label{section6}}
Deep CNNs have made a large impact in a broad range of application domains. Today, they are the first choice to solve many problems in computer vision, speech recognition and natural language processing. Taking computational advantage of working with small 2D and 3D patches, rather than the entire slice or volume, researchers in brain MR image analysis can train deep CNNs to obtain accurate segmentation algorithms. This success has received overwhelming acceptance by the community, in which shallow architectures were predominantly used. Currently, most CNN architectures have many layers, including additional normalisation layers, such as batch normalisation. Furthermore, each architecture is becoming increasingly more sophisticated, employing ideas from optimisation and probabilistic models.

About the architectures and their performances, the majority of the proposed works in Table \ref{tab:architecture_summary} used 2D networks. Although working with each model has its computational advantages and drawbacks, obtaining good generalisation requires an architecture with optimised layers, considering the class imbalance and selecting the best hyper-parameters and advanced training procedures. From the results shown in Table~\ref{tab:evaluation_summary}, methods with 2D CNN architecture with sufficient depth~\cite{pereira2016brain}, cascade~\cite{havaei2016brain}, ``short-cut connections" \cite{brosch2016deep} and parallel networks~\cite{moeskops2016automatic,de2015deep} showed top performance in their respective applications. After all, there is no universal architecture, and the ongoing research contributes to obtaining a model that can learn to provide a good representation of the underlying input image without suffering from significant over-fitting.

In the reviewed works, we observed that most of the proposed methods emphasised the shortcomings of working with deep CNN models. First, there is the computational requirement. Analysing, manipulating and processing each voxel in a volume is expensive computationally. The enormous amount of memory needed to store the extracted patches and a large amount of time required to process them constitute a difficulty. The deep learning software libraries used to implement layers of deep CNNs have either parallel or distributed frameworks, which help researchers to train their models in multi-core architectures or GPUs. Second, training CNN models for brain image analysis is hindered by the data imbalance problem, especially with small lesion or structure segmentation. For instance, in tumour or MS lesion segmentation, obtaining good generalisation is a challenge as most of the lesions are smaller than the entire volume. Two-phase training~\cite{havaei2016brain}, careful patch selection~\cite{pereira2016brain,moeskops2016automatic,lai2015deep} and loss functions~\cite{brosch2016deep, sudre2017generalised,fidon2017wassertein} are among the proposed strategies to overcome this problem.

Although we observed the success of CNNs; their full capacity has not yet been fully leveraged in brain MRI analysis. Networks can perform exceptionally when trained and tested on images with similar acquisition characteristics (e.g. resolution, intensity range), but when tested on sets with slight variations concerning the training set, they tend to fall behind traditional methods like FAST~\citep{FAST2001} and SPM~\citep{SPM5, SMP8}. This is the leading cause why traditional approaches are still being used in medical centres due to their robustness and adaptability~\citep{valverde2017automated}. Even though there is progress in domain adaptation techniques, more research in this sense is needed before being able to use resilient CNN-based software on medical centres.

It is becoming a widespread practice in the computer vision community to release source codes to the public. This practice helps to expedite the research in the field. The most commonly used deep learning libraries for MRI segmentation are Tensorflow\footnote{\url{http://www.tensorflow.org/}}, Theano\footnote{\url{http://deeplearning.net/software/theano/}}, Caffe\footnote{\url{http://caffe.berkeleyvision.org/}}, Keras\footnote{\url{https://keras.io/}} and PyTorch\footnote{\url{http://pytorch.org/}}. Also, open-source frameworks with implementation of major CNN approaches can be found online, such as NiftyNet\footnote{\url{http://www.niftynet.io/}} and DLTK\footnote{\url{http://dltk.github.io/}}. Another recommended practice is validating the model on different datasets. Few works~\cite{kleesiek2016deep, brosch2016deep,pereira2016brain,kamnitsas2016efficient,moeskops2016automatic,mehta2017brainsegnet} have reported their results on three or more different public datasets. This practice opens the door to design a robust model that can be applied to datasets of similar applications but with different types of MRI scanners, imaging modalities and numbers of training cases.

With the lack of training data, the poor spatial resolution of MR images and the need for a short prediction time, it has been impossible to train considerably deep CNNs. To train such networks, a considerable effort is needed in designing faster methods to perform convolutions. FFT algorithms~\cite{brosch2016deep} and quicker matrix multiplication methods \cite{lavin2015fast} have been used to improve the computation speed of CNNs, but there is yet room for improvement in the training algorithms of deep CNNs using variants of SGD~\cite{paine2013gpu} and their parallelised and distributed implementations. The new algorithms to come are expected not only to improve the performance of deep CNNs but also to be highly optimised, with less or no hyper-parameters, which constitute one of the major bottlenecks for most users to tune. 

\section*{Acknowledgments}

Jose Bernal and Kaisar Kushibar hold FI-DGR2017 grants from the Catalan Government with reference numbers 2017FI B00476 and 2017FI B00372, respectively. Daniel S. Asfaw holds an FI-DGR2016 grant from the Catalan Government. This work has been partially supported by La Fundació la Marató de TV3, by MPC UdG 2016/022 grant, by Retos de Investigació TIN2014-55710-R, TIN2015-73563-JIN and DPI2017-86696-R from the Ministerio de Ciencia y Tecnolog\'ia. The authors gratefully acknowledge the support of the NVIDIA Corporation with their donation of the Tesla K40 and the Titan X PASCAL GPUs used in this research.

\appendix

\bibliography{mybibfile}

\end{document}